%% file: main.tex
\documentclass[10pt,twocolumn,letterpaper]{article}

\usepackage[pagenumbers]{cvpr} 

\input{preamble}
\definecolor{cvprblue}{rgb}{0.21,0.49,0.74}
\usepackage[pagebackref,breaklinks,colorlinks,allcolors=cvprblue]{hyperref}
\usepackage{amsthm}
\usepackage{array}
\newtheorem{proposition}{Proposition}

\def\confName{CVPR}
\def\confYear{2027}

\newcommand{\proposal}{\textit{FedBARE}}

\title{FedBARE: Server-Side Recalibration for\\ Federated Open-Set Semi-Supervised Learning}

\author{You Rim Choi\\
Seoul National University\\
{\tt\small yrchoi@snu.ac.kr}
\and
Seojun Heo\\
Seoul National University\\
{\tt\small seodalzzz@snu.ac.kr}
\and
Subeom Park\\
Seoul National University\\
{\tt\small sbpark7@snu.ac.kr}
\and
Hyung-Sin Kim\\
Seoul National University\\
{\tt\small hyungkim@snu.ac.kr}
}

\begin{document}
\maketitle
\input{sec/0_abstract}    
\input{sec/1_intro_new}

\input{sec/2_related_work}

\input{sec/3_method}
\input{sec/4_experiment}
\input{sec/5_conclusion}

\input{sec/LLM_Usage_Statement}

{
    \small
    \bibliographystyle{ieeenat_fullname}
    \bibliography{main}
}
\input{sec/X_suppl}

\end{document}

%% file: preamble.tex
\usepackage{cuted} 

\usepackage{currfile} 

\usepackage{caption} 

%% file: sec/0_abstract.tex
\begin{abstract}
In federated learning, obtaining reliable client labels is costly.
A practical semi-supervised configuration therefore combines a small labeled server set with unlabeled client data, which may contain classes absent from the server labels.
This federated open-set semi-supervised learning (FOSSL) setting faces class-biased pseudo-labeling of out-of-distribution (OOD) samples and limited coverage of labeled negatives for OOD rejection.
We propose \proposal{} (Federated Balancing and Reference Supervision), which recalibrates a closed-set classifier and a one-vs-all (OVA) detector after each aggregation using only the server's labeled data.
Competitive-Response Balancing (CRB) penalizes imbalance across classifier outputs in their mean response to labeled samples of competing known classes.
Geometric Reference Supervision (GRS) constructs positive and negative feature references for each OVA head from the geometry of labeled features, with the negative-reference radius set by the nearest competing class.
Both losses update only their respective heads, require no external OOD data, and add no client computation or communication; clients use confidence-based pseudo-labeling and soft detector consistency.
Across 12 CIFAR-10/100 settings with IID and non-IID clients, \proposal{} improves accuracy and AUROC over our reproduction of FedOpenMatch by 6.1 and 22.0 percentage points on average.
On SVHN it improves AUROC over the strongest baseline at comparable accuracy.
Ablations and pseudo-label analyses support the complementary benefits of CRB and GRS.
\end{abstract}

%% file: sec/1_intro_new.tex
\section{Introduction}
\label{sec:introduction}

\begin{figure}[t]
  \centering
  \includegraphics[width=\linewidth]{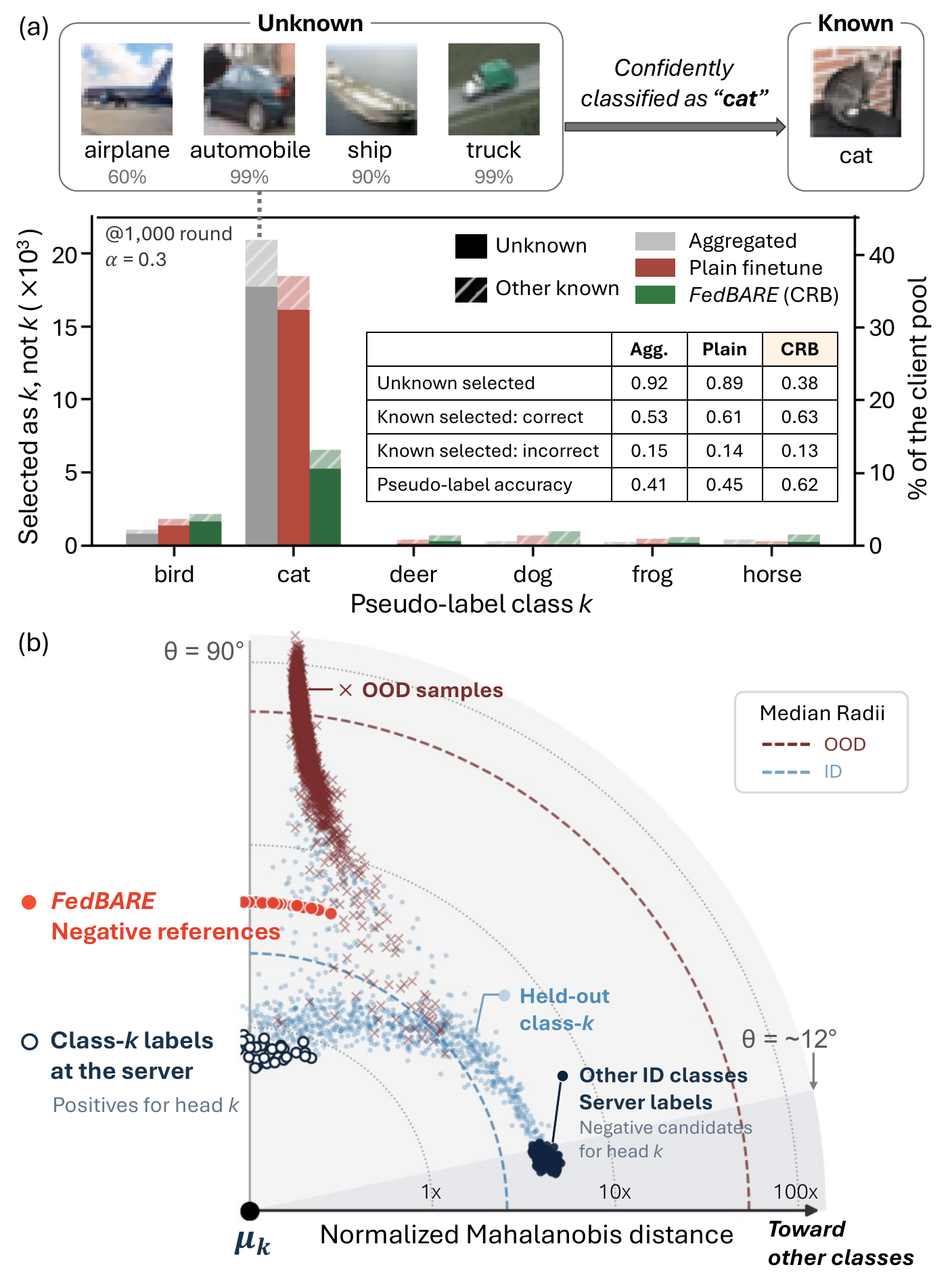}
    \caption{\textbf{Two challenges motivating \proposal{}.}
    Single-run diagnostics on CIFAR-10 with 6 ID classes, 4 OOD classes, and 40 labeled samples per ID class at the server, under a non-IID Dirichlet partition with \(\alpha=0.3\).
    (a) Unknown images misclassified as \textit{cat} (top) and incorrect pseudo-label counts by assigned class (bottom). Bars and the inset compare the aggregated model (Aggregated), supervised server fine-tuning (Plain), and fine-tuning with CRB. Percentages below the example images indicate the fraction of images in each unknown class selected with \textit{cat} as their pseudo-label (confidence \(\ge 0.95\)) under Aggregated.
    (b) Limited coverage of labeled negatives for OVA head \(k\), with GRS negative references.
    }
    \vspace{-2ex}
  \label{fig:fossl_challenges}
\end{figure}

In federated learning, participating devices or institutions, called clients, train a model on locally held data. 
A server aggregates their updates into a global model and sends it back for the next round~\cite{mcmahan2017fedavg}. 
When labeling images at every client is impractical, the server can maintain a small labeled set while clients learn from unlabeled images~\cite{diao2022semifl,jeong2021fedmatch}.
We study federated open-set semi-supervised learning (FOSSL) in this labels-at-server setting~\cite{liu2026fedopenmatch}. 
Client images may belong to known classes represented in the server's labeled set (in-distribution, ID) or to unknown classes outside it (out-of-distribution, OOD). 
Clients also differ in which classes they hold and in how many unknowns they contain, so their data are non-IID.
The goal is to learn from these distributed images while recognizing known classes and rejecting unknowns.

Unknown images pose a problem even before inference. 
A client can use the global model's confident prediction as a training target, or pseudo-label. 
For an unknown image, every known-class label is wrong, yet the model can still assign one with high confidence.
Fig.~\ref{fig:fossl_challenges}(a) illustrates the problem in a CIFAR-10 run with six known animal classes and four unknown vehicle classes. 
Images from distinct unknown classes can receive the same known-class label, \textit{cat}. 
Evaluated over all 100 client pools after plain server fine-tuning, the broadcast classifier would select 89\% of unknown vehicle images, but only 61\% of known animal images with their correct labels. 
Such concentration is known from centralized semi-supervised learning~\cite{han2022pseudolabel}; here it persists right after the server has used its verified labels, and the resulting targets guide client updates that have no labeled supervision to correct them.

Rejecting unknowns poses a second problem.
A class-wise one-vs-all (OVA) detector tests whether an image belongs to the class predicted by the classifier~\cite{saito2021ovanet,saito2021openmatch}.
For OVA head \(k\), server labels provide positives from class \(k\) and negatives from the other known classes.
These negatives cluster in directions toward those classes, while unknown images can lie in directions with little labeled coverage (Fig.~\ref{fig:fossl_challenges}(b)).

What more can the server learn from its labels before the next broadcast?
A labeled image of class \(j\) carries evidence for both problems.
The \textit{logit} of each competing class \(k\ne j\) measures how strongly that output responds to a verified non-member, and the image's \textit{feature vector} marks a valid negative location for OVA head \(k\).
SemiFL and FedOpenMatch already fine-tune the global model after aggregation~\cite{diao2022semifl,liu2026fedopenmatch}, and FedOpenMatch adjusts logits for positive–negative imbalance within each OVA head~\cite{liu2026fedopenmatch}.
Going further, \textbf{\proposal{} (Federated Balancing and Reference Supervision)} turns both signals into objectives for the same server update: it (1) balances how each classifier output responds to the other known classes and (2) extends class-wise rejection supervision beyond the observed negatives.

\begin{itemize} [leftmargin=*]
    \item \textbf{Competitive-Response Balancing (CRB).}
Cross-entropy favors the correct class on each labeled image, but an additive preference for one classifier output also raises its logit on images from other known classes.
CRB measures this preference as each output's mean logit over verified non-members, averaged within each competing class and weighted equally across classes, and penalizes the variance of these means across outputs.
Unlike post-hoc corrections estimated from off-target logits~\cite{tang2026intrinsic}, the penalty acts before the classifier generates the next round's client pseudo-labels.
In the single-round diagnostic of Fig.~\ref{fig:fossl_challenges}(a), CRB sharply reduces the concentration on \textit{cat}, so fewer unknown images pass the confidence threshold, while correct selection of known images is preserved.

\item \textbf{Geometric Reference Supervision (GRS).}
For each OVA head, GRS pairs two kinds of references in whitened feature space.
Negative references lie along uniformly drawn directions on a shell whose radius encloses the Gaussian reference region of the nearest competing class (Fig.~\ref{fig:fossl_challenges}(b)), extending rejection supervision to directions without labeled negatives.
Positive references follow directions observed within the target class, in the radial band between its labeled features and the Gaussian reference scale, so that the expanded rejection coverage does not erode the head's positive supervision.
Unlike synthesized outlier features for OOD detection~\cite{du2022vos,tao2023npos,karzanov2026gcos}, each GRS negative carries a rejection target for one head only, so it remains valid supervision even where its location is plausible for another known class.
\end{itemize}

Both objectives use only the server's labeled data and update their respective heads. 
They require no external unknown-class examples and add no client computation or communication to the classifier–detector pipeline. 
Clients select pseudo-labels using classifier confidence alone and train the detector only through soft open-set consistency regularization (SOCR) across augmented views~\cite{saito2021openmatch}. 
Together with the supervised server losses, these updates shape subsequent feature learning and client pseudo-labeling.

Across 12 CIFAR-10/100 settings with IID and non-IID clients, \proposal{} improves known-class accuracy by 6.1 percentage points (pp) and AUROC by 22.0~pp on average over reproduced FedOpenMatch, with maximum gains of 13.0 and 40.2~pp, respectively.
On SVHN with ResNet-18, it also improves AUROC over the strongest baseline while maintaining comparable accuracy.
Component ablations and pseudo-label analyses support the complementary benefits of CRB and GRS.

%% file: sec/2_related_work.tex
\section{Related Work}
\label{sec:related_work}

\subsection{Federated Open-Set Semi-Supervised Learning}

Labels-at-server federated semi-supervised learning (SSL) places a small labeled dataset at the server while clients hold only unlabeled data ~\cite{jeong2021fedmatch,diao2022semifl}. SemiFL ~\cite{diao2022semifl} is a representative method that alternates supervised server fine-tuning with client training using pseudo-labels generated by the global model.

When client data contain classes absent from the server labels, centralized open-set SSL (OSSL) methods ~\cite{wang2024scomatch,wallin2024prosub,fan2023ssb} can be adapted by applying supervised losses at the server and unlabeled objectives at clients. However, their original objectives were designed for joint optimization rather than separate training stages. FedOpenMatch ~\cite{liu2026fedopenmatch} targets the same labels-at-server FOSSL setting as ours. It combines a classifier with a class-wise OVA detector using logit adjustment for binary imbalance, gradient stopping to reduce feature interference, and logit consistency regularization (LCR). However, its server objectives do not directly correct class-biased classifier responses. This can lead to incorrect OOD pseudo-labeling during client training. Moreover, verified negatives for each OVA head come only from labeled samples of other ID classes. Client-side OVA pseudo-negatives are selected based on the detector's own predictions, which may reinforce existing errors rather than reliably correct them. 

~\proposal{} retains the classifier-OVA architecture and alternating server and client training of FedOpenMatch while addressing both limitations through server-side recalibration using only labeled ID data.

\subsection{Class Bias in Pseudo-Label Selection}

Prior SSL methods adjust pseudo-label selection through class-adaptive thresholds~\cite{zhang2021flexmatch,wang2023freematch} and distribution alignment ~\cite{berthelot2020remixmatch}. In class-mismatch SSL, Han et al.~\cite{han2022pseudolabel} rebalance high-confidence selections after identifying concentrated OOD pseudo-labels.
In FOSSL, these statistics are estimated from ID/OOD mixtures and reflect client distributions that differ from the server's labeled distribution.
Logit adjustment~\cite{menon2021logit_adjustment} uses class priors, while FedOpenMatch applies it to the binary imbalance within OVA heads~\cite{liu2026fedopenmatch}.
Tang et al.~\cite{tang2026intrinsic} estimate a post-hoc correction from off-target logits.
CRB instead computes each classifier output's mean logit on labeled examples of other known classes, with equal weight per class.
It penalizes imbalance across outputs using a loss normalized by overall logit variance during server training.
The correction thus uses only labeled server examples and is incorporated into the broadcast model before the next round's pseudo-label selection.

\subsection{Rejection Supervision without OOD Examples}
Server labels provide each OVA head with negatives only at observed feature locations of other known classes (Fig.~\ref{fig:fossl_challenges}(b)).
OSSL methods exploit unlabeled data through cross-view consistency~\cite{saito2021openmatch}, which we retain, and through pseudo-negative mining~\cite{fan2023ssb}, unknown-class modeling~\cite{li2023iomatch,wang2024scomatch}, or score-distribution estimation~\cite{wallin2024prosub}.
In our federated adaptations, these unlabeled objectives operate on heterogeneous client pools without concurrent labeled supervision.
Outlier Exposure~\cite{hendrycks2019oe} requires real auxiliary outliers, while VOS~\cite{du2022vos}, NPOS~\cite{tao2023npos}, and GCOS~\cite{karzanov2026gcos} synthesize outliers from class-conditional Gaussians, boundary features, or Mahalanobis-quantile shells for global ID/OOD separation.
GRS instead provides class-wise positive and negative reference supervision for OVA detector heads using labeled feature geometry~\cite{lee2018mahalanobis, chen2010shrinkage}.

%% file: sec/3_method.tex
\section{\proposal{}}
\label{sec:method}

\begin{figure*}[t]
  \centering
  \includegraphics[width=\textwidth]{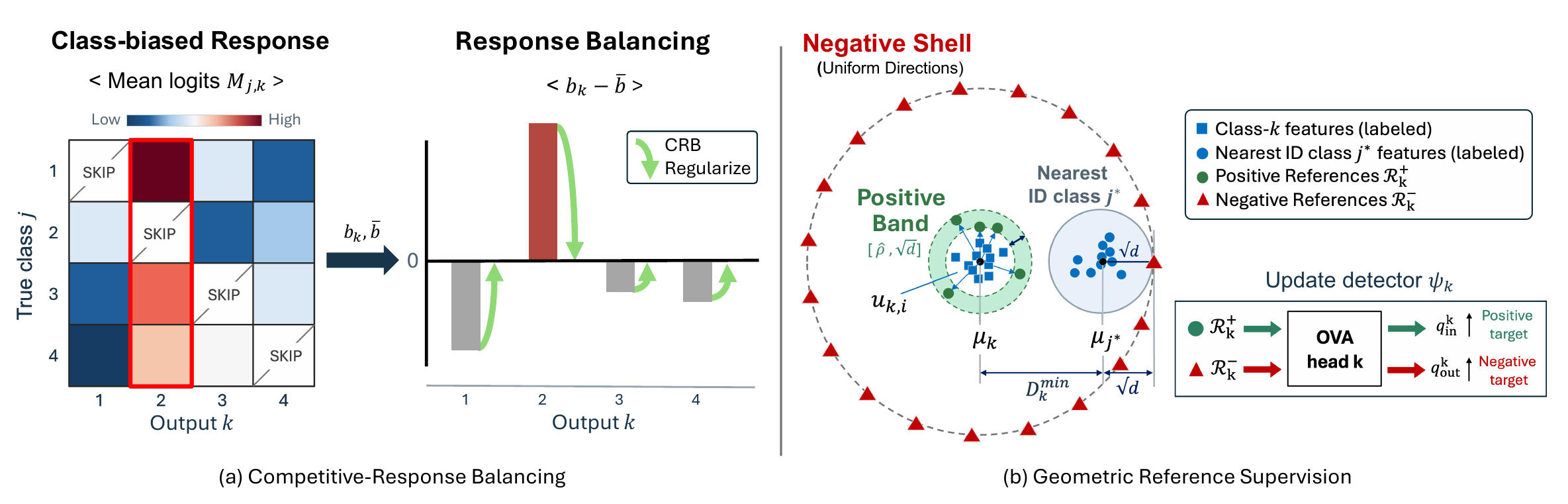}
  \caption{\textbf{Server-side recalibration in \proposal{}.}
  (a) CRB penalizes imbalance in the off-diagonal column means \(b_k\) of \(M\). Bars show centered responses \(b_k-\bar b\). The rightmost diagram shows the balancing target, \(b_k-\bar b=0\).
  (b) GRS pairs positive references \(\mathcal R_k^+\) along labeled residual directions with negative references \(\mathcal R_k^-\) on an isotropic shell of radius \(D_k^{\min}+\sqrt d\).
  Each reference supervises its corresponding OVA head. Geometry is illustrated in whitened space with \(\hat\rho<\sqrt d\).}
  \label{fig:fedgr}
\end{figure*}

\subsection{Overview}
\label{sec:FedGR_overview}

The server holds a small class-balanced labeled ID set \(\mathcal D_s=\{(x_i,y_i)\}_{i=1}^{N_s}\), with \(K\) classes.
Clients hold unlabeled ID/OOD mixtures with non-IID data distributions across clients.
Our model consists of a feature extractor \(f_\theta\), a closed-set classifier \(g_\phi\), and an OVA detector parameterized by \(\psi\), whose \(K\) heads share one MLP.
For feature \(h=f_\theta(x)\in\mathbb R^d\), the classifier produces logits \(z=g_\phi(h)\in\mathbb R^K\).
Detector head \(k\) outputs binary probabilities \(q^k(h)=(q^k_{\rm in}(h),q^k_{\rm out}(h))\), denoting class \(k\) and not class \(k\), respectively~\cite{saito2021ovanet,saito2021openmatch}.
We define \(P(x)=\operatorname{softmax}(g_\phi(f_\theta(x)))\) and \(Q^k(x)=q^k(f_\theta(x))\) for image-level predictions.
After supervised initialization, training alternates client updates, FedAvg aggregation~\cite{mcmahan2017fedavg}, and server fine-tuning.

\paragraph{Server learning.}
The server uses cross-entropy \(\mathcal L_{\rm CE}\) for the classifier and supervised OVA loss \(\mathcal L_{\rm OVA}\) for the detector.
Let \(\mathbb E_{\mathcal B}\) denote the sample average over batch \(\mathcal B\).
For a labeled batch \(\mathcal B_s\), 

\begin{equation}
\mathcal L_{\rm OVA}
=-\mathbb E_{\mathcal B_s}\!\Big[
\log Q^y_{\rm in}(x)
+\sum_{k\in\mathcal N(x,y)}\log Q^k_{\rm out}(x)
\Big],
\label{eq:supervised_ova}
\end{equation}
where \(\mathcal N(x,y)\) combines the detector's hardest negative head, as in~\cite{saito2021ovanet,saito2021openmatch}, with the classifier's strongest competing class. Both selections exclude \(y\), and a shared index is counted once.
\proposal{} adds CRB (\S\ref{sec:crb}) to balance classifier responses across classes and GRS (\S\ref{sec:grs}) to extend OVA supervision beyond labeled features (Fig.~\ref{fig:fedgr}).

\paragraph{Client learning.}
Let \(x_i^{w_1},x_i^{w_2}\) denote two weak views and \(x_i^s\) a strong view. 
Each client caches soft targets \(p_i=\operatorname{softmax}(z^{\rm global}(x_i^{w_1})/T)\) and confidence masks \(m_i=\mathbf 1[\max_k p_{ik}\ge\tau]\)
from the broadcast model.
Both remain fixed for the round.
The classifier matches cached targets on strong views~\cite{sohn2020fixmatch}, while the detector uses SOCR on weak views~\cite{saito2021openmatch}.
For an unlabeled batch \(\mathcal B_u\),

\begin{equation}
\begin{aligned}
&\mathcal L_{\rm client}
=\mathbb E_{\mathcal B_u}\!\Big[
m_i\,\ell_{\rm cons}(p_i,P(x_i^s))\\
&\quad+\lambda_{\rm SOCR}\sum_{k=1}^K
\|Q^k(x_i^{w_1})-Q^k(x_i^{w_2})\|_2^2\Big],
\end{aligned}
\label{eq:client_objective}
\end{equation}
where \(\ell_{\rm cons}(p,P)=-\sum_k p_k\log P_k\), \(T\) is the temperature, and \(\tau\) is the confidence threshold.
Both SOCR predictions use the current client model and receive gradients.
Selection uses classifier confidence alone; clients apply no entropy minimization or pseudo-negative mining.

\subsection{Competitive-Response Balancing (CRB)}
\label{sec:crb}

CRB uses classifier responses to other ID classes as a label-derived proxy for class preference, targeting the selection bias in Fig.~\ref{fig:fossl_challenges}(a).
For a labeled batch containing at least two classes, let \(\mathcal J\) be the classes present and \(n_j\) their sample counts.
Using logits \(z_i=g_\phi(\operatorname{sg}(f_\theta(x_i)))\), where \(\operatorname{sg}\) stops gradients, define

\begin{equation}
M_{jk}=\frac{1}{n_j}\sum_{i:y_i=j}z_{ik},
\qquad
b_k=\frac{1}{|\mathcal J\setminus\{k\}|}
\sum_{\substack{j\in\mathcal J\\j\ne k}}M_{jk}.
\label{eq:competitive_response}
\end{equation}
In Fig.~\ref{fig:fedgr}(a), rows of \(M\) represent true classes and columns represent classifier outputs.
Averaging each column without its diagonal gives the competitive response \(b_k\), with equal weight for each competing class.

\begin{proposition}[Competitive responses and prediction invariance]
\label{prop:crb_shifts}
For a fixed labeled batch, competitive responses have the following properties.

\noindent
\textup{(i)} Under shifts \(z'_k(x)=z_k(x)+\delta_k\), where
\(\delta\in\mathbb R^K\) is shared across inputs, let
\(P'(x)=\operatorname{softmax}(z'(x))\). Then
\begin{equation}
\begin{aligned}
b'_k-b'_\ell
&=(b_k-b_\ell)+\delta_k-\delta_\ell,\\
\log\frac{P'_k(x)}{P'_\ell(x)}
&=\log\frac{P_k(x)}{P_\ell(x)}+\delta_k-\delta_\ell.
\end{aligned}
\label{eq:crb_shift_odds}
\end{equation}
\textup{(ii)} For arbitrary \(c_i\in\mathbb R\), the shift
\(z'_{ik}=z_{ik}+c_i\) leaves every \(P(x_i)\) unchanged.
If all \(K\) classes occur in the batch, there exist such shifts
for which all competitive responses \(b'_k\) are equal.
\end{proposition}
Part (i) links additive class preferences to competitive-response contrasts and prediction odds.
Let \(v_i\) denote the input to the classifier's final linear layer, which has no bias, and write \(z_i=Wv_i\).
Then \(b_k=w_k^\top\bar v_{\neg k}\), where \(\bar v_{\neg k}\) averages the feature means of other classes in the batch with equal weight.
Thus \(b_k\) measures the classifier's average response to other-class features through its learned weights.
Part (ii) shows that response balance can change without affecting softmax predictions; we therefore evaluate selection rates and pseudo-label accuracy.

With \(\bar b=K^{-1}\sum_k b_k\), the bars labeled \(b_k-\bar b\) in Fig.~\ref{fig:fedgr}(a) visualize response imbalance.
CRB penalizes their dispersion,
\begin{equation}
\mathcal L_{\rm CRB}=
\frac{\frac{1}{K}\sum_{k=1}^K(b_k-\bar b)^2}
{V+\epsilon},
\qquad V=\operatorname{Var}_{i,k}(z_{ik}),
\label{eq:crb_loss}
\end{equation}
where \(V\) is the variance over all batch logits and \(\epsilon>0\) ensures numerical stability.
The \emph{Balancing target} in Fig.~\ref{fig:fedgr}(a) illustrates \(b_k-\bar b=0\) for every output.
We differentiate both numerator and denominator to respect the loss's invariance to uniform logit rescaling when \(\epsilon=0\).
The denominator also contributes gradients to true-class logits excluded from \(b_k\).
We detach features to recalibrate classifier responses without directly reshaping the shared representation; supervised server losses continue to train the feature extractor.
The proof and gradient analysis are provided in the supplementary material.

\subsection{Geometric Reference Supervision (GRS)}
\label{sec:grs}
 
GRS addresses the limited negative coverage in Fig.~\ref{fig:fossl_challenges}(b) using labeled feature geometry.
It pairs negative references \(\mathcal R_k^-\) that broaden rejection coverage with positive references \(\mathcal R_k^+\) that preserve class-\(k\) supervision along observed within-class directions (Fig.~\ref{fig:fedgr}(b)).

\paragraph{Feature geometry.}
From \(d\)-dimensional server features, we estimate class means \(\mu_k\) and a shared within-class covariance \(S\).
We stabilize the covariance using Oracle Approximating Shrinkage (OAS)~\cite{chen2010shrinkage},
\begin{equation}
\widehat\Sigma=(1-\gamma)S+
\gamma\frac{\operatorname{tr}(S)}{d}I+\eta I.
\label{eq:feature_geometry}
\end{equation}
Here \(\gamma\) is determined by OAS from labeled residuals, and \(\eta>0\) is a small fixed diagonal jitter for numerical stability.
Let \(L\) be the lower-triangular Cholesky factor satisfying \(\widehat\Sigma=LL^\top\).
We use the Mahalanobis distance~\cite{lee2018mahalanobis}, \(d_M(h,\mu_k)=\|L^{-1}(h-\mu_k)\|_2\).
For \(H\sim\mathcal N(\mu_k,\widehat\Sigma)\), \(\mathbb E[d_M(H,\mu_k)^2]=d\), giving the root-mean-square radius \(\sqrt d\) as our reference scale.

\paragraph{Positive references \(\mathcal R_k^+\).}
Positive references retain observed within-class directions and provide supervision across the radial band between labeled features and the Gaussian reference scale \(\sqrt d\).
Let \(\hat\rho\) denote an upper quantile of labeled distances \(d_M(h_i,\mu_{y_i})\), pooled across classes.
For a labeled feature \(h_i\) of class \(k\) with nonzero residual, we use its unit whitened direction \(u_{k,i}\) and sample a Mahalanobis radius \(r\),

\begin{equation}
\begin{gathered}
u_{k,i}=\frac{L^{-1}(h_i-\mu_k)}{d_M(h_i,\mu_k)},
\quad h^+_{k,i}=\mu_k+rLu_{k,i},\\
r\sim\mathcal U\big(\min(\hat\rho,\sqrt d),\max(\hat\rho,\sqrt d)\big).
\end{gathered}
\label{eq:positive_references}
\end{equation}
The sampled offsets are mapped back to the original feature space through \(L\), yielding the green references in Fig.~\ref{fig:fedgr}(b).
We retain a candidate in \(\mathcal R_k^+\) only if its nearest class mean under \(d_M\) is \(\mu_k\), and assign it a positive target for head \(k\). 
They complement the expanded rejection supervision by reinforcing positive predictions of head \(k\) along observed within-class directions.

\paragraph{Negative references \(\mathcal R_k^-\).}
We construct negative references to supervise rejection across directions at a distance anchored to the nearest other ID class.
For head \(k\), let \(j^*\) denote such a class under \(d_M\). We sample directions uniformly in whitened space and set the radius using the distance to \(\mu_{j^*}\) and the Gaussian reference spread,

\begin{equation}
\begin{alignedat}{2}
D_k^{\min}&=\min_{j\ne k}d_M(\mu_k,\mu_j),
&\quad r_k&=D_k^{\min}+\sqrt d,\\
h^-_k&=\mu_k+r_kLu,
&u&\sim\operatorname{Unif}(\mathbb S^{d-1}).
\end{alignedat}
\label{eq:negative_references}
\end{equation}
Here \(\mathbb S^{d-1}\) is the unit sphere in \(\mathbb R^d\), and \(L\) maps the sampled offsets back to the original feature space.
The term \(D_k^{\min}\) measures the distance to the nearest competing mean, while \(\sqrt d\) accounts for that class's Gaussian reference spread.
The orange triangles in Fig.~\ref{fig:fedgr}(b) illustrate the resulting shell of Mahalanobis radius \(r_k\).
Each sample enters \(\mathcal R_k^-\) and receives a negative target only for head \(k\).
These references need not lie outside every ID class and are not assigned global OOD labels.

\begin{proposition}[Geometry of the negative-reference radius]
\label{prop:grs_geometry}
Fix an OVA head \(k\), and let \(j^*\in\arg\min_{j\ne k}d_M(\mu_k,\mu_j)\) be a nearest other ID class.
For any radius \(R\ge0\), define the reference ellipsoid
\[
\mathcal B_j(R)=
\{h\in\mathbb R^d:d_M(h,\mu_j)\le R\}.
\]
The radius \(r_k=D_k^{\min}+\sqrt d\) satisfies
\begin{equation}
r_k=\min_{j\ne k}\sup_{h\in\mathcal B_j(\sqrt d)}
d_M(h,\mu_k).
\label{eq:grs_envelope}
\end{equation}
Let \(X\sim\mathcal N(\mu_{j^*},\widehat\Sigma)\) follow the Gaussian reference model of class \(j^*\), and define its root-mean-square distance from \(\mu_k\) as \(s_k=\sqrt{\mathbb E[d_M(X,\mu_k)^2]}\).
Then
\begin{equation}
s_k=\sqrt{(D_k^{\min})^2+d},
\qquad s_k\le r_k\le\sqrt2\,s_k.
\label{eq:grs_locality}
\end{equation}
\end{proposition}
Thus \(r_k\) is the smallest Mahalanobis radius centered at \(\mu_k\) that encloses the reference ellipsoid of class \(j^*\).
Moreover, it is at most \(\sqrt2\) times that class's root-mean-square distance from \(\mu_k\).
The proof is provided in the supplementary material.

\paragraph{Reference loss.}
GRS combines positive and negative reference supervision,
\begin{equation}
\begin{aligned}
\mathcal L_{\rm GRS}
={}&-\sum_{k:\,|\mathcal R_k^+|>0}
\frac{\sum_{h\in\mathcal R_k^+}\log q^k_{\rm in}(h)}
{|\mathcal R_k^+|}\\
&-\sum_{k:\,|\mathcal R_k^-|>0}
\frac{\sum_{h\in\mathcal R_k^-}\log q^k_{\rm out}(h)}
{|\mathcal R_k^-|}.
\end{aligned}
\label{eq:geometric_loss}
\end{equation}
Positive and negative terms are averaged separately for each head, with empty reference sets omitted.
Reference features are detached, so GRS updates only the detector parameters \(\psi\).
Because the heads share a detector MLP, an update driven by one head can also affect other heads' predictions.

\subsection{Server-Side Recalibration}
\label{sec:inference_scope}
 
After aggregation, the server fine-tunes with

\begin{equation}
\begin{aligned}
\mathcal L_{\rm server}={}&\mathcal L_{\rm CE}
+\lambda_{\rm OVA}\mathcal L_{\rm OVA}\\
&+\lambda_{\rm CRB}\mathcal L_{\rm CRB}
+\lambda_{\rm GRS}\mathcal L_{\rm GRS}.
\end{aligned}
\label{eq:server_objective}
\end{equation}
Before each server fine-tuning epoch that uses GRS, we re-estimate \((\mu_k,\widehat\Sigma,\hat\rho)\) from the current labeled features. The geometry remains fixed within the epoch, while references \(\mathcal R_k^\pm\) are resampled at every optimization step.
CRB adds constraints across classifier outputs, while GRS adds detector targets beyond labeled features.
Both use only server labels and update their respective branches without directly updating the feature extractor.
Supervised server losses and client objectives continue to train the backbone.
The branch updates change subsequent supervised feature gradients. In the next round, the recalibrated classifier supplies cached targets and confidence masks, while the recalibrated detector influences client feature learning through SOCR. 
The added objectives require no external OOD data or additional client computation or communication.
Implementation details are provided in the supplementary material.

At inference, the classifier selects an ID class and its OVA head supplies the OOD score,
\begin{equation}
\hat y(x)=\arg\max_k z_k(x),\qquad
a(x)=Q^{\hat y(x)}_{\rm out}(x).
\label{eq:inference_score}
\end{equation}
Larger scores indicate stronger evidence for rejection.

%% file: sec/4_experiment.tex
\input{Tables/main_cifar}

\section{Experiments}
\label{sec:experiments}

\subsection{Experimental Setup}
\label{sec:experimental_setup}

\noindent
\textbf{Datasets and backbones.}
We evaluate on CIFAR-10/100~\cite{krizhevsky2009learning} and SVHN~\cite{netzer2011reading}.
CIFAR-10 uses 6 known animal classes and 4 unknown vehicle classes and CIFAR-100 uses 55/45 and 80/20 known/unknown class splits.
SVHN uses digits 0--5 as known and 6--9 as unknown.
We use 40 server labels per known class on CIFAR, with an additional 20-label setting for CIFAR-10, and 20 on SVHN.
The remaining training images form the client pools.
We use WRN-28-2~\cite{zagoruyko2016wide} for CIFAR.
For SVHN, we use ResNet-18~\cite{he2016deep} adapted for \(32\times32\) images.

\noindent
\textbf{Federated setting and implementation.}
We simulate 100 clients with 10 randomly selected per round, using IID or Dirichlet partitions (\(\alpha\in\{0.3,0.1\}\)) over known and unknown classes.
After 500 epochs of supervised server initialization, we train for 2,500 communication rounds on CIFAR and 1,000 on SVHN.
Each round consists of five local training epochs, FedAvg aggregation~\cite{mcmahan2017fedavg}, and three server fine-tuning epochs.
We implement federated training using the Flower framework~\cite{beutel2020flower}.
Further implementation details, baseline reproduction details, and additional results are provided in the supplementary material.

\noindent
\textbf{Baselines.}
We compare with SemiFL~\cite{diao2022semifl}, FedOpenMatch~\cite{liu2026fedopenmatch}, and federated adaptations of FixMatch~\cite{sohn2020fixmatch}, SCOMatch~\cite{wang2024scomatch}, ProSub~\cite{wallin2024prosub}, and SSB~\cite{fan2023ssb}.
These adaptations apply supervised losses at the server and unlabeled objectives at clients.
Class splits, label budgets, client partitions, and backbones are matched across methods.

\noindent
\textbf{Evaluation protocol.}
We report top-1 accuracy on known-class test images and OOD detection AUROC.
Seen OOD classes are present in the unlabeled client pools.
Unseen OOD evaluation uses CIFAR-10/100, SVHN, LSUN~\cite{yu2015lsun}, ImageNet-30~\cite{hendrycks2019using}, and Gaussian noise, excluding each model's training dataset.
Overall AUROC equally weights seen-OOD AUROC and the mean AUROC across unseen datasets.
We use maximum softmax probability (MSP)~\cite{hendrycks2017baseline} for SemiFL and FedFixMatch.
Other baselines follow the OOD scoring rules in their original implementations, while \proposal{} uses scores from its OVA detector.
All methods are evaluated with their EMA models.

\subsection{Main Results}
\label{sec:main_results}

\input{Tables/svhn}

Table~\ref{tab:main_results} compares known-class accuracy and overall AUROC across the CIFAR settings.
\proposal{} achieves the highest mean accuracy and AUROC in most settings, and no baseline outperforms it on both metrics in any setting.
On CIFAR-10, its AUROC margin over the strongest baseline is larger at \(\alpha=0.1\) than under IID partitioning, highlighting its effectiveness under strong client heterogeneity.

Compared with reproduced FedOpenMatch~\cite{liu2026fedopenmatch}, \proposal{} improves AUROC across all settings, with gains of 26.7--40.2 percentage points (pp) on CIFAR-10 and 6.8--26.0~pp on CIFAR-100.
The upper end comes from the 55/45 split with \(\alpha=0.1\), where FedOpenMatch's AUROC stays at chance in all three runs (6.8--14.4~pp otherwise).
Known-class accuracy also improves except on CIFAR-10 with 20 labels per class and \(\alpha=0.1\).
Here, \proposal{} has 3.2~pp lower mean accuracy but 26.7~pp higher AUROC, with a smaller accuracy standard deviation across runs (1.2 versus 2.3~pp).
We also observe pronounced AUROC fluctuations during training in several FedOpenMatch runs.
These results highlight the importance of evaluating unknown rejection alongside known-class classification.

SemiFL~\cite{diao2022semifl} exhibits training collapse in some open-set runs despite supervised server updates.
This illustrates that support for unlabeled clients alone does not ensure robust learning from open-set data.
The federated adaptations of SCOMatch and ProSub are less competitive overall.
SCOMatch couples selected OOD pseudo-examples with known-class supervision, while ProSub uses labeled examples to anchor its ID subspace and score-distribution estimates~\cite{wang2024scomatch,wallin2024prosub}.
Both methods were originally designed for centralized settings where labeled and unlabeled data are jointly available, rather than for separate server and client training.

FedSSB is generally the strongest detection baseline.
It shares the classifier-OVA structure of our base protocol but retains entropy minimization and pseudo-negative mining.
Compared with FedSSB, \proposal{} improves accuracy by 0.5--8.6~pp across all CIFAR settings and achieves higher AUROC in all non-IID settings.
Section~\ref{sec:ablation} isolates the contributions of CRB and GRS under fixed client training.

To extend our evaluation beyond CIFAR and WRN-28-2, Table~\ref{tab:svhn_results} reports results on SVHN using ResNet-18.
FedSSB remains the most accurate baseline.
\proposal{} is within 0.2~pp of its accuracy or better and exceeds the best baseline AUROC by 1.7--10.0~pp.
These results support its applicability to a different dataset and backbone with a larger feature dimension (512 versus 128).

\subsection{Ablation Studies}
\label{sec:ablation}

\input{Tables/ablation}

\paragraph{Effect of CRB and GRS.}
We compare four variants that share the client objective (Eq.~\ref{eq:client_objective}) and differ only in the server loss.
Base fine-tunes with \(\mathcal L_{\rm CE}+\lambda_{\rm OVA}\mathcal L_{\rm OVA}\), and +CRB, +GRS, and \proposal{} add the corresponding terms of Eq.~\ref{eq:server_objective} (Table~\ref{tab:ablation}).
Base already exceeds the reproduced FedOpenMatch in AUROC on CIFAR-10 (Table~\ref{tab:main_results}), so we credit CRB and GRS only with the gains over Base.
CRB alone improves accuracy in every setting (0.1--1.9~pp) but changes AUROC inconsistently (\(-0.7\) to \(+4.2\)~pp).
GRS alone improves AUROC in every setting (0.9--5.5~pp) and accuracy in three.
Combining them improves accuracy by 0.4--2.3~pp and AUROC by 2.7--7.2~pp over Base, with the highest accuracy in all four settings (one tie) and the highest AUROC in three.
The exception is CIFAR-100 with 55 known classes, where GRS alone reaches higher AUROC but lower accuracy.

\begin{figure}[t]
\centering
\includegraphics[width=\linewidth]{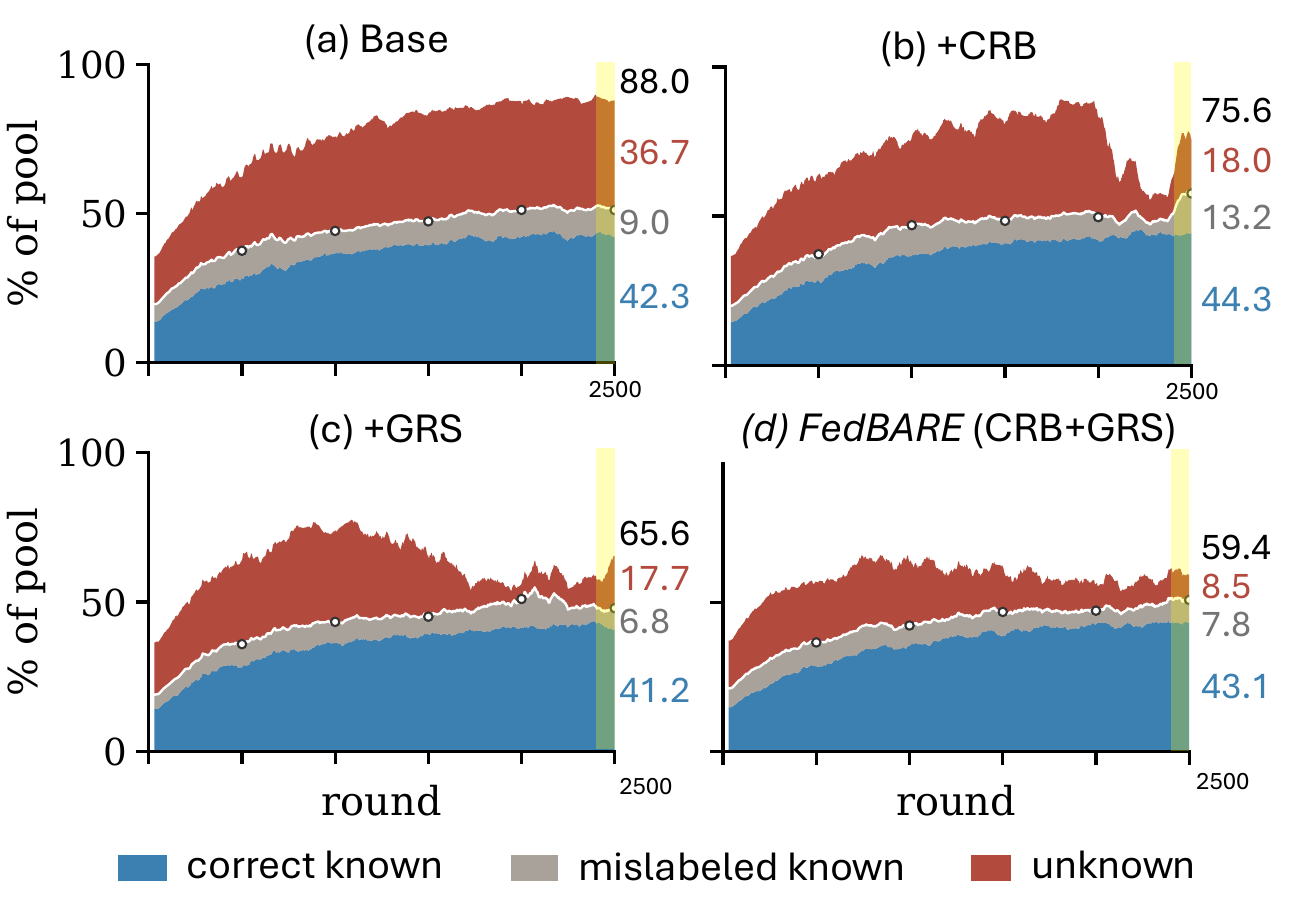}
\caption{Composition of the confidence-gated client pools over training (CIFAR-10, 40 labels per class, Dirichlet $\alpha{=}0.3$). Each band is the share of the pool that passes the gate: correct known (blue), mislabeled known (grey), unknown (red). All shares are logged every round during client training. The values beside each plot report the mean proportions of each category over the final 75 rounds (yellow shaded).}
\vspace{-1ex}
\label{fig:gateflow}
\end{figure}

\begin{figure}[t]
\centering
\includegraphics[width=\linewidth]{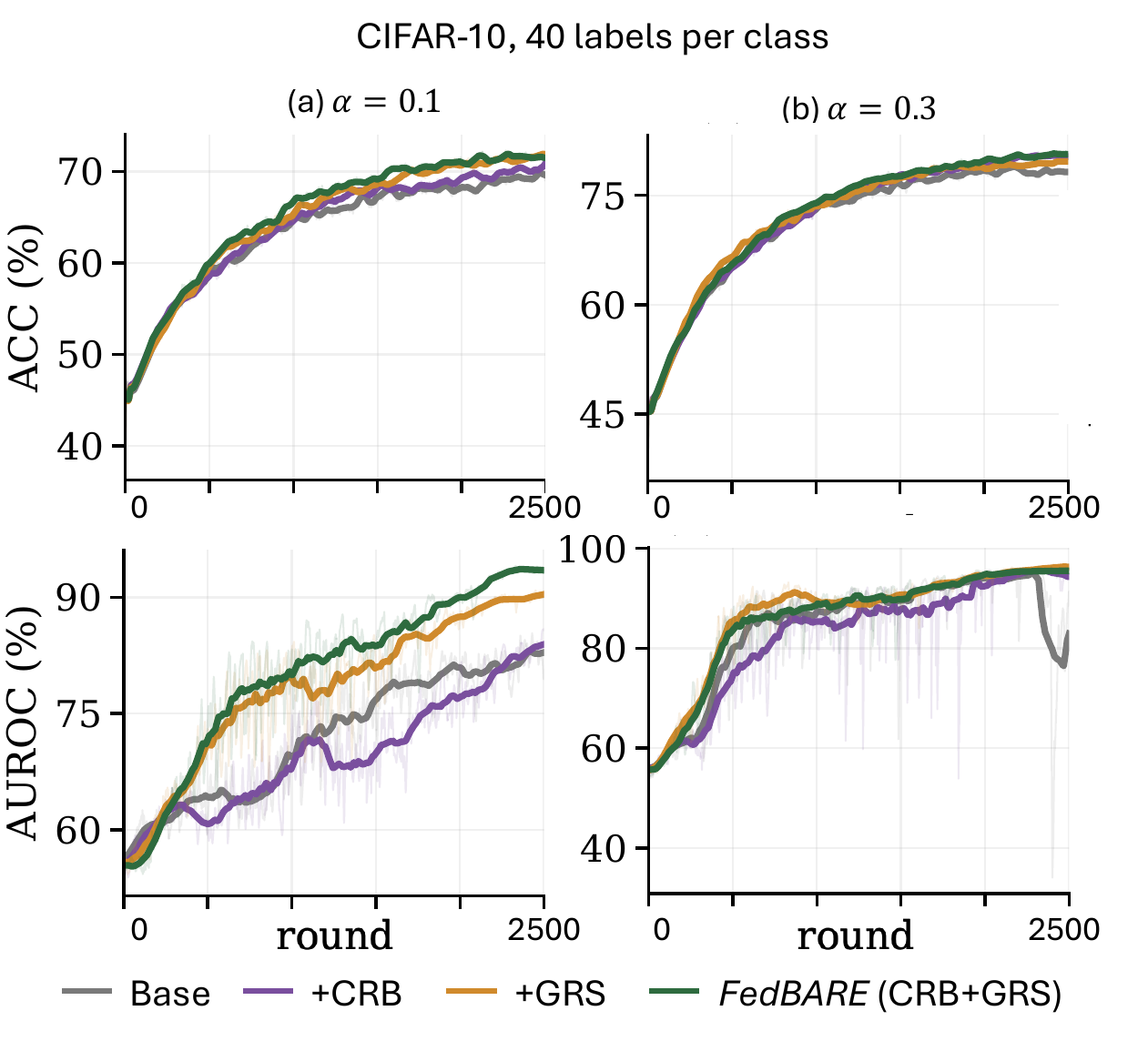}
\caption{Learning curves of the four variants (CIFAR-10, 40 labels per class, $\alpha{=}0.1,0.3$). The top row shows closed-set accuracy and the bottom row shows seen-OOD AUROC. Thin lines are per-round values, while thick lines show moving averages.}
\vspace{-1ex}
\label{fig:curves}
\end{figure}

\paragraph{Interaction of CRB and GRS.}
Figure~\ref{fig:gateflow} decomposes the confidence-selected client samples at \(\alpha=0.3\) into correct known, mislabeled known, and unknown shares of the client pool.
Each objective alone roughly halves the selected-unknown share of Base (36.7\% to 18.0\% with CRB and 17.7\% with GRS), and together they reduce it to 8.5\%.
Each also offsets a side effect of the other.
CRB alone admits more mislabeled known samples and GRS alone slightly fewer correct ones, while the combination shows neither.
Pseudo-label accuracy among selected samples thus rises from 48.1\% (Base) to 58.6\% (+CRB), 62.7\% (+GRS), and 72.6\% (\proposal{}).
This pattern is consistent with the objectives correcting different errors.
CRB rebalances the classifier outputs that select client targets, removing selections driven by a preferred class, but leaves each OVA head with the same labeled negatives.
GRS updates only the detector and reaches the classifier indirectly, through SOCR and the shared features, so its accuracy gains are consistent with training on cleaner pseudo-labels.
Figure~\ref{fig:curves} shows a similar dependence in seen-OOD AUROC, while accuracy curves remain close.
At \(\alpha=0.1\), CRB alone ends near Base, whereas \proposal{} ends above GRS alone, suggesting that balanced responses aid detection mainly when the heads also receive reference supervision.
At \(\alpha=0.3\), GRS alone and \proposal{} finish close, while Base declines late in the illustrated run.

%% file: Tables/main_cifar.tex
\begin{table*}[t]
\centering
\footnotesize
\def\std#1{{\fontsize{7pt}{8pt}\selectfont ($\pm$#1)}}
\def\mrow#1{\raisebox{-\dimexpr\arraystretch\normalbaselineskip/2\relax}[0pt][0pt]{#1}}
\def\thickhline{\noalign{\hrule height .08em}}
\setlength{\tabcolsep}{3pt}
\settowidth{\dimen0}{\std{00.0}}%
\settowidth{\dimen2}{FedOpenMatch}\divide\dimen2 by 2
\ifdim\dimen2>\dimen0 \dimen0=\dimen2 \fi
\settowidth{\dimen2}{\textbf{\proposal{}}}\divide\dimen2 by 2
\ifdim\dimen2>\dimen0 \dimen0=\dimen2 \fi
\edef\colw{\the\dimen0}%
\settowidth{\dimen2}{Client Partitioning}%
\edef\firstw{\the\dimexpr 0.85\dimen2\relax}%
\def\hd#1{\makebox[\colw]{#1}}%
\long\def\tablebody{%
\toprule
\mrow{\begin{tabular}{@{}c@{}}Client\\Partitioning\end{tabular}}
  & \multicolumn{2}{c|}{SemiFL}
  & \multicolumn{2}{c|}{FedFixMatch}
  & \multicolumn{2}{c|}{FedSCOMatch}
  & \multicolumn{2}{c|}{FedProSub}
  & \multicolumn{2}{c|}{FedSSB}
  & \multicolumn{2}{c|}{FedOpenMatch}
  & \multicolumn{2}{c}{\textbf{\proposal{}}} \\
\cline{2-15}
  \makebox[\firstw]{} & \hd{Acc.} & \hd{AUC} & \hd{Acc.} & \hd{AUC} & \hd{Acc.} & \hd{AUC} & \hd{Acc.} & \hd{AUC} & \hd{Acc.} & \hd{AUC} & \hd{Acc.} & \hd{AUC} & \hd{Acc.} & \hd{AUC} \\
\thickhline
\multicolumn{15}{c}{\textbf{CIFAR-10 6/4, 20 Labels/Class}} \\  
\thickhline
\mrow{IID}
   & 38.9 & 45.3 & 76.6 & 40.1 & 69.5 & 47.3 & 39.2 & 59.3 & \underline{81.4} & \underline{91.5} & 72.9 & 62.2 & \textbf{85.9} & \textbf{92.8} \\
   & \std{9.7} & \std{7.4} & \std{11.2} & \std{5.9} & \std{0.7} & \std{3.6} & \std{19.3} & \std{15.3} & \std{3.3} & \std{3.4} & \std{8.4} & \std{7.2} & \std{0.8} & \std{3.4} \\
\mrow{$\alpha$=0.3}
   & 27.4 & 49.5 & 65.0 & 41.5 & 60.4 & 46.6 & 31.9 & 48.0 & 64.4 & \underline{79.2} & \underline{65.7} & 47.5 & \textbf{71.0} & \textbf{87.7} \\
   & \std{18.5} & \std{5.8} & \std{1.1} & \std{16.5} & \std{2.3} & \std{4.0} & \std{4.9} & \std{4.0} & \std{1.1} & \std{8.6} & \std{10.3} & \std{1.7} & \std{3.7} & \std{3.0} \\
\mrow{$\alpha$=0.1}
   & 16.7 & 49.9 & 53.9 & 49.4 & 52.3 & 53.9 & 25.9 & 37.6 & 51.6 & \underline{71.2} & \textbf{63.4} & 51.4 & \underline{60.2} & \textbf{78.1} \\
   & \std{0.0} & \std{3.7} & \std{0.4} & \std{2.9} & \std{4.3} & \std{2.6} & \std{2.2} & \std{9.4} & \std{2.8} & \std{1.3} & \std{2.3} & \std{5.8} & \std{1.2} & \std{4.1} \\
\thickhline
\multicolumn{15}{c}{\textbf{CIFAR-10 6/4, 40 Labels/Class}} \\  
\thickhline
\mrow{IID}
   & 50.3 & 38.1 & 86.3 & 56.9 & 75.2 & 52.4 & 62.3 & 75.0 & \underline{87.7} & \textbf{91.7} & 76.2 & 59.6 & \textbf{89.0} & \underline{90.0} \\
   & \std{31.0} & \std{14.2} & \std{1.1} & \std{6.2} & \std{0.7} & \std{2.7} & \std{3.2} & \std{1.6} & \std{1.1} & \std{1.4} & \std{10.5} & \std{12.5} & \std{1.0} & \std{0.9} \\
\mrow{$\alpha$=0.3}
   & 51.9 & 51.5 & 74.5 & 51.6 & 65.8 & 54.2 & 51.5 & 56.7 & \underline{74.8} & \underline{87.8} & 74.1 & 64.4 & \textbf{79.3} & \textbf{91.8} \\
   & \std{30.7} & \std{9.6} & \std{2.4} & \std{5.1} & \std{1.3} & \std{3.5} & \std{2.1} & \std{0.6} & \std{3.0} & \std{3.4} & \std{3.1} & \std{5.6} & \std{1.1} & \std{0.4} \\
\mrow{$\alpha$=0.1}
   & 16.7 & 44.1 & 65.5 & 56.0 & 59.6 & 55.2 & 43.1 & 43.5 & 65.5 & \underline{74.9} & \underline{68.2} & 52.2 & \textbf{69.5} & \textbf{84.0} \\
   & \std{0.0} & \std{11.8} & \std{1.0} & \std{2.4} & \std{1.7} & \std{3.3} & \std{3.4} & \std{8.0} & \std{2.2} & \std{1.8} & \std{4.4} & \std{4.3} & \std{1.7} & \std{4.8} \\
\thickhline
\multicolumn{15}{c}{\textbf{CIFAR-100 55/45, 40 Labels/Class}} \\  
\thickhline
\mrow{IID}
   & 63.1 & 68.3 & 66.4 & 68.0 & 66.6 & 65.1 & 60.5 & 75.9 & \underline{69.0} & \underline{76.3} & 62.7 & 72.3 & \textbf{70.4} & \textbf{79.1} \\
   & \std{0.4} & \std{0.6} & \std{0.8} & \std{1.1} & \std{1.7} & \std{0.9} & \std{0.5} & \std{0.3} & \std{0.7} & \std{2.7} & \std{0.7} & \std{1.2} & \std{0.7} & \std{0.2} \\
\mrow{$\alpha$=0.3}
   & 61.2 & 67.6 & 63.6 & 65.3 & 64.2 & 61.5 & 58.7 & \underline{75.3} & \underline{66.7} & 74.8 & 62.4 & 64.3 & \textbf{67.3} & \textbf{78.7} \\
   & \std{0.5} & \std{0.3} & \std{1.0} & \std{2.8} & \std{0.8} & \std{1.3} & \std{1.4} & \std{0.9} & \std{0.9} & \std{2.3} & \std{1.8} & \std{14.0} & \std{1.0} & \std{1.3} \\
\mrow{$\alpha$=0.1}
   & 59.5 & 65.2 & 60.3 & 65.7 & 61.8 & 64.4 & 57.8 & 71.7 & \underline{63.5} & \underline{73.0} & 59.3 & 50.0 & \textbf{64.0} & \textbf{76.0} \\
   & \std{1.1} & \std{1.7} & \std{1.2} & \std{0.7} & \std{1.1} & \std{1.1} & \std{1.0} & \std{1.0} & \std{0.6} & \std{1.8} & \std{1.6} & \std{0.0} & \std{0.4} & \std{2.0} \\
\thickhline
\multicolumn{15}{c}{\textbf{CIFAR-100 80/20, 40 Labels/Class}} \\  
\thickhline
\mrow{IID}
   & 56.8 & 66.1 & 59.7 & 62.2 & 58.0 & 57.8 & 49.7 & 73.8 & \underline{60.6} & \underline{75.1} & 53.4 & 63.8 & \textbf{63.5} & \textbf{75.5} \\
   & \std{1.0} & \std{1.2} & \std{0.3} & \std{0.6} & \std{0.5} & \std{2.4} & \std{1.0} & \std{1.9} & \std{0.4} & \std{2.3} & \std{0.5} & \std{3.1} & \std{0.6} & \std{1.1} \\
\mrow{$\alpha$=0.3}
   & 54.1 & 64.7 & 56.3 & 60.5 & 56.5 & 57.4 & 48.1 & 71.7 & \underline{57.3} & \underline{72.6} & 53.0 & 64.8 & \textbf{60.3} & \textbf{73.7} \\
   & \std{0.5} & \std{0.2} & \std{0.5} & \std{0.9} & \std{0.9} & \std{2.5} & \std{0.1} & \std{0.4} & \std{0.2} & \std{1.8} & \std{0.7} & \std{1.9} & \std{0.6} & \std{2.8} \\
\mrow{$\alpha$=0.1}
   & 52.4 & 61.2 & 52.8 & 60.7 & 53.7 & 58.6 & 46.5 & 70.1 & \underline{54.5} & \underline{70.6} & 52.2 & 62.3 & \textbf{56.9} & \textbf{70.9} \\
   & \std{0.6} & \std{1.8} & \std{0.6} & \std{2.8} & \std{0.1} & \std{1.8} & \std{0.4} & \std{2.2} & \std{1.0} & \std{1.3} & \std{0.3} & \std{2.5} & \std{1.1} & \std{1.3} \\
\bottomrule
}%
\sbox0{\begin{tabular}{c|cc|cc|cc|cc|cc|cc|cc}\tablebody\end{tabular}}%
\setlength{\tabcolsep}{\dimexpr\tabcolsep+(\textwidth-\wd0)/30\relax}%
\ifdim\tabcolsep<1pt \setlength{\tabcolsep}{1pt}\fi
\caption{Known-class accuracy and overall AUROC (AUC) on CIFAR-10/100, reported in \%. Each cell shows the mean (top) and standard deviation (bottom) over three independent runs. The best result in each row is in \textbf{bold} and the second best is \underline{underlined}, compared separately for Acc.\ and AUC.}
\label{tab:main_results}
\begin{tabular}{c|cc|cc|cc|cc|cc|cc|cc}\tablebody\end{tabular}
\end{table*}

%% file: Tables/svhn.tex
\begin{table*}[t]
\centering
\footnotesize
\def\std#1{{\fontsize{7pt}{8pt}\selectfont ($\pm$#1)}}
\def\mrow#1{\raisebox{-\dimexpr\arraystretch\normalbaselineskip/2\relax}[0pt][0pt]{#1}}
\def\thickhline{\noalign{\hrule height .08em}}
\setlength{\tabcolsep}{3pt}
\settowidth{\dimen0}{\std{00.0}}%
\settowidth{\dimen2}{FedOpenMatch}\divide\dimen2 by 2
\ifdim\dimen2>\dimen0 \dimen0=\dimen2 \fi
\settowidth{\dimen2}{\textbf{\proposal{}}}\divide\dimen2 by 2
\ifdim\dimen2>\dimen0 \dimen0=\dimen2 \fi
\edef\colw{\the\dimen0}%
\settowidth{\dimen2}{Client Partitioning}%
\edef\firstw{\the\dimexpr 0.95\dimen2\relax}%
\def\hd#1{\makebox[\colw]{#1}}%
\long\def\tablebody{%
\toprule
\mrow{\begin{tabular}{@{}c@{}}Client\\Partitioning\end{tabular}}
  & \multicolumn{2}{c|}{SemiFL}
  & \multicolumn{2}{c|}{FedFixMatch}
  & \multicolumn{2}{c|}{FedSCOMatch}
  & \multicolumn{2}{c|}{FedProSub}
  & \multicolumn{2}{c|}{FedSSB}
  & \multicolumn{2}{c|}{FedOpenMatch}
  & \multicolumn{2}{c}{\textbf{\proposal{}}} \\
\cline{2-15}
  \makebox[\firstw]{} & \hd{Acc.} & \hd{AUC} & \hd{Acc.} & \hd{AUC} & \hd{Acc.} & \hd{AUC} & \hd{Acc.} & \hd{AUC} & \hd{Acc.} & \hd{AUC} & \hd{Acc.} & \hd{AUC} & \hd{Acc.} & \hd{AUC} \\
\thickhline
\multicolumn{15}{c}{\textbf{SVHN 6/4, 20 Labels/Class}} \\  
\thickhline
\mrow{IID}
   & 94.1 & \underline{80.4} & 94.9 & 76.4 & 95.3 & 60.7 & 61.1 & 73.1 & \underline{95.8} & 76.6 & 71.1 & 73.9 & \textbf{97.1} & \textbf{82.1} \\
   & \std{2.4} & \std{2.4} & \std{1.4} & \std{2.8} & \std{0.4} & \std{5.6} & \std{7.0} & \std{7.4} & \std{0.0} & \std{1.9} & \std{0.7} & \std{1.9} & \std{1.1} & \std{1.3} \\
\mrow{$\alpha$=0.3}
   & 90.4 & \underline{82.8} & 83.8 & 74.3 & 90.9 & 64.5 & 80.5 & 73.8 & \textbf{95.1} & 77.9 & 71.8 & 67.0 & \underline{94.9} & \textbf{92.8} \\
   & \std{0.8} & \std{4.5} & \std{14.0} & \std{2.6} & \std{1.6} & \std{5.6} & \std{3.6} & \std{7.1} & \std{0.2} & \std{0.9} & \std{2.4} & \std{12.6} & \std{2.1} & \std{4.5} \\
\mrow{$\alpha$=0.1}
   & 84.8 & 79.3 & 59.4 & 69.7 & 82.7 & 59.0 & 75.0 & 70.9 & \underline{88.1} & \underline{80.2} & 71.5 & 67.5 & \textbf{88.2} & \textbf{86.8} \\
   & \std{4.1} & \std{4.6} & \std{4.6} & \std{2.2} & \std{3.6} & \std{3.7} & \std{3.4} & \std{0.5} & \std{6.5} & \std{3.3} & \std{4.9} & \std{12.1} & \std{4.1} & \std{5.4} \\
\bottomrule
}%
\sbox0{\begin{tabular}{c|cc|cc|cc|cc|cc|cc|cc}\tablebody\end{tabular}}%
\setlength{\tabcolsep}{\dimexpr\tabcolsep+(\textwidth-\wd0)/30\relax}%
\ifdim\tabcolsep<1pt \setlength{\tabcolsep}{1pt}\fi
\caption{Known-class accuracy and overall AUROC (AUC) on SVHN (6 known / 4 unknown classes, 20 labels per known class), reported in \%. Mean (top) and standard deviation (bottom) over three independent runs; best in \textbf{bold}, second best \underline{underlined}.}
\label{tab:svhn_results}
\begin{tabular}{c|cc|cc|cc|cc|cc|cc|cc}\tablebody\end{tabular}
\end{table*}

%% file: Tables/ablation.tex
\begin{table}[t]
\centering
\scriptsize
\setlength{\tabcolsep}{4pt}
\renewcommand{\arraystretch}{1}
\caption{Ablation of CRB and GRS under Dirichlet $\alpha{=}0.1$. Single runs with a shared seed, so values differ from the means in Table~\ref{tab:main_results}.}
\label{tab:ablation}
\resizebox{\columnwidth}{!}{%
\begin{tabular}{l cc cc cc}
\toprule
\multicolumn{7}{l}{\textit{CIFAR-10} (6/4)} \\
 & & & \multicolumn{2}{c}{40 labels / class}
 & \multicolumn{2}{c}{20 labels / class} \\
\cmidrule(lr){4-5}\cmidrule(lr){6-7}
Variant & CRB & GRS & ACC & AUROC & ACC & AUROC \\
\midrule
Base        &            &            & 69.5          & 80.5             & 57.8             & 78.6             \\
+CRB        & \checkmark &            & 70.7          & 84.7             & \underline{59.7} & 78.9             \\
+GRS        &            & \checkmark & \textbf{71.3} & \underline{86.0} & 59.0             & \underline{79.5} \\
\proposal{} & \checkmark & \checkmark & \textbf{71.3} & \textbf{87.7}    & \textbf{60.1}    & \textbf{81.3}    \\
\midrule
\multicolumn{7}{l}{\textit{CIFAR-100} (40 labels / class)} \\
 & & & \multicolumn{2}{c}{55/45}
 & \multicolumn{2}{c}{80/20} \\
\cmidrule(lr){4-5}\cmidrule(lr){6-7}
Variant & CRB & GRS & ACC & AUROC & ACC & AUROC \\
\midrule
Base        &            &            & 63.3             & 73.2             & 55.9             & 66.6             \\
+CRB        & \checkmark &            & \underline{63.5} & 72.5             & 56.0             & 65.9             \\
+GRS        &            & \checkmark & 62.7             & \textbf{78.4}    & \underline{56.7} & \underline{68.6} \\
\proposal{} & \checkmark & \checkmark & \textbf{63.7}    & \underline{76.8} & \textbf{57.7}    & \textbf{70.1}    \\
\bottomrule
\end{tabular}%
}
\end{table}

%% file: sec/5_conclusion.tex
\section{Conclusion}
\label{sec:conclusion}

We presented \proposal{}, a server-side recalibration method for federated open-set semi-supervised learning with labels only at the server.
Using only these labels and no added client cost, CRB balances classifier responses to labeled non-members before the next round of pseudo-labeling, and GRS supervises each OVA head with positive and negative references built from labeled feature geometry.
Across CIFAR-10/100 and SVHN, \proposal{} achieves the highest known-class accuracy in most settings and the highest AUROC in every non-IID setting, and ablations indicate that the two objectives correct different pseudo-labeling errors.
One limitation is that GRS relies on a shared-covariance Gaussian geometry estimated from few labels, and its references are geometric targets rather than samples of real unknown classes.

%% file: sec/LLM_Usage_Statement.tex

\section*{LLM Usage Statement}
We used ChatGPT (OpenAI) and Claude (Anthropic) in this work. For writing and editing, the tools generated and revised English text based on drafts and guidance provided by the authors, who reviewed, edited, and approved the resulting text. For related work and citations, the tools generated draft text and citation candidates, which the authors reviewed, verified, and approved. For code and implementation, the authors established the codebase structure, within which the tools generated substantial portions of the additional implementation and experiment code. The authors reviewed and verified the generated code. For experiments, the tools assisted in refining the authors' experiment designs, execution procedures, and analyses of results. For figures and visual materials, the tools generated drafts, which the authors reviewed, edited, and approved. For research ideas, the tools assisted in discussing and refining ideas, hypotheses, and claims proposed by the authors. We provided the full manuscript or major parts of it to OpenAI and Anthropic models to obtain reviewer-style feedback. No LLM or AI tools were used for synthetic data generation, dataset creation, or annotation. The authors take full responsibility for all submitted content, regardless of any LLM or AI-tool use.

%% file: sec/X_suppl.tex
\clearpage
\setcounter{page}{1}
\maketitlesupplementary

\newif\ifFBaddnormcontrol
\FBaddnormcontrolfalse
\newif\ifFBaddreferencecontrol
\FBaddreferencecontrolfalse
\newif\ifFBaddreproduction
\FBaddreproductionfalse
\newif\ifFBaddsemiflcontrol
\FBaddsemiflcontrolfalse

\providecommand{\tbd}{\textit{TBD}}
\providecommand{\FBcrbprop}{\ref{prop:crb_shifts}}
\providecommand{\FBgrsprop}{\ref{prop:grs_geometry}}

\newcounter{FBalgorithm}
\renewcommand{\theFBalgorithm}{S\arabic{FBalgorithm}}
\providecommand{\theHFBalgorithm}{supp.algorithm.\arabic{FBalgorithm}}
\providecommand{\FBalgorithmautorefname}{Algorithm}
\newcounter{FBalgline}

\appendix
\setcounter{table}{0}
\setcounter{figure}{0}
\setcounter{equation}{0}

\renewcommand{\thetable}{S\arabic{table}}
\renewcommand{\thefigure}{S\arabic{figure}}
\renewcommand{\theequation}{S\arabic{equation}}

\noindent
This supplement first provides the proofs and gradient analysis supporting the main-paper propositions in Section~\ref{app:analysis}.
Sections~\ref{app:crb_experiments} and~\ref{app:grs_experiments} present controlled analyses of Competitive-Response Balancing (CRB) and Geometric Reference Supervision (GRS).
Section~\ref{app:implementation} describes the implementation and training procedure of \proposal{}.
Section~\ref{app:baseline_implementation} documents the baseline implementations.
Section~\ref{app:additional_results} presents expanded evaluation.

\input{sec/supple/A_proof_and_analysis}
\input{sec/supple/B_CRB}
\input{sec/supple/C_GRS}
\input{sec/supple/D_implementation}

\input{sec/supple/E_baseline}
\input{sec/supple/F_expanded_eval}


%% file: sec/supple/A_proof_and_analysis.tex
\section{Proofs and Analysis}
\label{app:analysis}

Sections~\ref{app:crb_analysis} and~\ref{app:grs_analysis} provide the CRB and GRS proofs and analyses. Section~\ref{app:indirect_gradients} describes their indirect feature-gradient paths. 
Their empirical effects are evaluated in Sections~\ref{app:crb_experiments} and~\ref{app:grs_experiments}.

\subsection{CRB: Proofs and Analysis}
\label{app:crb_analysis}

\subsubsection{Competitive-Response Statistics}

Fix a labeled batch \(\{(x_i,y_i)\}_{i=1}^n\) with \(K\ge2\) classifier outputs. 
Let \(\mathcal J\) be the represented class set, with \(|\mathcal J|\ge2\), and \(n_j\) the number of class-\(j\) examples. 
Writing \(z_{ik}\) for the output-\(k\) logit on \(x_i\), define
\begin{equation}
\begin{alignedat}{2}
\mathcal J_{\neg k} &=\mathcal J\setminus\{k\}, &\quad M_{jk} &=\frac{1}{n_j}\sum_{i:y_i=j}z_{ik},\\
b_k &=\frac{1}{|\mathcal J_{\neg k}|} \sum_{j\in\mathcal J_{\neg k}}M_{jk}, &\quad \bar b &=\frac{1}{K}\sum_{k=1}^{K}b_k.
\end{alignedat}
\label{eq:supp_response_definition}
\end{equation}
Here \(M_{jk}\) is the class-conditional mean logit for \(j\in\mathcal J\) and \(b_k\) is the competitive response.

For the bias-free final layer \(z_i=Wv_i\), let \(v_i=v(x_i)\) denote its input and \(w_k^\top\) row \(k\) of \(W\). 
Writing \(\bar v_j=n_j^{-1}\sum_{i:y_i=j}v_i\), we have
\begin{equation}
b_k=w_k^\top\bar v_{\neg k},\qquad \bar v_{\neg k} =\frac{1}{|\mathcal J_{\neg k}|} \sum_{j\in\mathcal J_{\neg k}}\bar v_j.
\end{equation}

\subsubsection{Class-Dependent Shifts and Prediction Odds}
\label{app:crb_shift_odds}

This proves part~(i) of Proposition~\FBcrbprop{} in the main paper. 
Let \(z'_k(x)=z_k(x)+\delta_k\), where \(\delta\in\mathbb R^K\) is shared across inputs. Class-wise averaging gives
\begin{equation}
M'_{jk} =\frac{1}{n_j}\sum_{i:y_i=j}(z_{ik}+\delta_k) =M_{jk}+\delta_k,
\end{equation}
and hence \(b'_k=b_k+\delta_k\). Therefore
\begin{equation}
b'_k-b'_\ell=(b_k-b_\ell)+\delta_k-\delta_\ell.
\end{equation}
Writing \(P(x)=\operatorname{softmax}(z(x))\) and \(P'(x)=\operatorname{softmax}(z'(x))\), we obtain
\begin{equation}
\begin{aligned}
\frac{P'_k(x)}{P'_\ell(x)} &=\frac{\exp(z_k(x)+\delta_k)} {\exp(z_\ell(x)+\delta_\ell)}\\
&=\frac{P_k(x)}{P_\ell(x)} \exp(\delta_k-\delta_\ell).
\end{aligned}
\end{equation}
Taking logarithms proves the odds identity, which holds for every input, including unknown-class inputs. The response identity concerns the fixed labeled batch. This is a logit-space sensitivity statement and does not assume an output-bias parameter.

\subsubsection{Prediction-Invariant Shifts}
\label{app:crb_invariance}

This proves part~(ii) of Proposition~\FBcrbprop{}. Let \(z'_i=z_i+c_i\mathbf 1\), where \(c_i\in\mathbb R\) and \(\mathbf 1\in\mathbb R^K\) is the all-ones vector. Then
\begin{equation}
\operatorname{softmax}(z'_i) =\operatorname{softmax}(z_i).
\end{equation}
Thus the batch softmax probabilities and predicted classes are unchanged.

For the conditional equalization claim, assume that all \(K\) classes occur in the fixed batch. This assumption is used only for the construction; CRB training does not require full class coverage. Choose
\begin{equation}
c_i=(K-1)(b_{y_i}-\bar b).
\end{equation}
Since each competing class contributes equally,
\begin{equation}
\begin{aligned}
b'_k &=b_k+\frac{1}{K-1} \sum_{j\ne k}(K-1)(b_j-\bar b)\\
&=b_k-(b_k-\bar b)=\bar b.
\end{aligned}
\end{equation}
The numerator, and hence the CRB loss for \(\epsilon>0\), is zero while batch predictions are preserved.

Now hold the mapping \(v(x)\) fixed and vary only \(W\). For any vector \(u\) of the same dimension as \(v(x)\), the update \(W\mapsto W+\mathbf 1u^\top\) adds \(u^\top v(x)\) to every logit of any input \(x\). Thus all softmax predictions, cached soft targets, and confidence masks remain unchanged, while
\begin{equation}
b'_k=b_k+u^\top\bar v_{\neg k}.
\label{eq:supp_row_shift}
\end{equation}
Under the same full-class assumption, if the class means \(\{\bar v_j\}_{j=1}^K\) are linearly independent, there exists \(u\) satisfying \(u^\top\bar v_j=(K-1)(b_j-\bar b)\) for every \(j\). Along \(W_t=W+t\mathbf 1u^\top\), \(0\le t\le1\), the mean response remains \(\bar b(t)=\bar b\). Writing \(N(t)=K^{-1}\sum_k(b_k(t)-\bar b)^2\), we obtain
\begin{equation}
\begin{aligned}
b_k(t)&=\bar b+(1-t)(b_k-\bar b),\\
N(t)&=(1-t)^2N(0).
\end{aligned}
\label{eq:supp_weight_shift_path}
\end{equation}
The path equalizes the responses while preserving every softmax prediction. Cross-entropy is constant, but the variance denominator and weight regularization can change; monotone decrease of the normalized CRB loss is not guaranteed.

Equalizing \(b_k\) alone therefore does not guarantee improved pseudo-label selection. With GRS disabled, Section~\ref{app:crb_experiments} compares CRB with post-hoc balancing and a row-centered variant. The row-centered variant computes both CRB terms from
\begin{equation}
\begin{aligned}
\widetilde z_{ik} &=z_{ik}-\frac{1}{K}\sum_{\ell=1}^K z_{i\ell}\\
&=\log P_k(x_i) -\frac{1}{K}\sum_{\ell=1}^K\log P_\ell(x_i).
\end{aligned}
\end{equation}
This removes common-shift sensitivity. For fixed labeled inputs, the objective depends only on their softmax probability vectors and cannot change while all of those vectors remain unchanged.

\subsubsection{CRB Gradients and Scale Invariance}
\label{app:crb_gradient}
\label{app:crb_scale}

For the raw-logit CRB objective, write \(\mathcal L_{\rm CRB}=N/(V+\epsilon)\), where \(N=K^{-1}\sum_k(b_k-\bar b)^2\), \(\bar z=(nK)^{-1}\sum_{i,k}z_{ik}\), and \(V=(nK)^{-1}\sum_{i,k}(z_{ik}-\bar z)^2\). Differentiating both numerator and denominator gives
\begin{equation}
\begin{aligned}
\frac{\partial\mathcal L_{\rm CRB}}{\partial z_{ik}} ={}&\frac{2\mathbf 1[y_i\ne k](b_k-\bar b)} {K n_{y_i}|\mathcal J_{\neg k}|(V+\epsilon)}\\
&-\frac{2\mathcal L_{\rm CRB}(z_{ik}-\bar z)} {nK(V+\epsilon)}.
\end{aligned}
\label{eq:crb_full_gradient}
\end{equation}
Here \(\mathbf 1[\cdot]\) denotes the indicator function. For a true-class logit \(k=y_i\), the numerator contribution is zero, but the denominator contribution can remain nonzero.

For \(\epsilon=0\) and \(V(z)>0\), both \(N\) and \(V\) scale quadratically, yielding
\begin{equation}
\begin{aligned}
\mathcal L_{\rm CRB}(sz)&=\mathcal L_{\rm CRB}(z), \qquad s\ne0,\\
\sum_{i,k}z_{ik} \frac{\partial\mathcal L_{\rm CRB}}{\partial z_{ik}}&=0.
\end{aligned}
\label{eq:supp_scale_invariance}
\end{equation}
Detaching \(V\) preserves the forward value but changes the latter inner product to \(2\mathcal L_{\rm CRB}\), removing this cancellation along the uniform scaling direction. For \(\epsilon>0\), scale invariance holds approximately when both \(V(z)\) and \(s^2V(z)\) are much larger than \(\epsilon\).

\subsection{GRS: Proofs and Analysis}
\label{app:grs_analysis}

\subsubsection{Negative-Reference Geometry}
\label{app:grs_geometry}

This proves Proposition~\FBgrsprop{} in the main paper. Fix fitted means \(\mu_k\in\mathbb R^d\) and the covariance \(\widehat\Sigma=LL^\top\) from main-paper Eq.~\eqref{eq:feature_geometry}. Here \(d\) is the backbone feature dimension and \(L\) is the Cholesky factor of \(\widehat\Sigma\). The OAS-shrunk part is positive semidefinite, and the jitter \(\eta>0\) makes \(\widehat\Sigma\) positive definite. Recall
\begin{equation}
\begin{aligned}
d_M(h,\mu)&=\|L^{-1}(h-\mu)\|_2,\\
D_k^{\min}&=\min_{j\ne k}d_M(\mu_k,\mu_j).
\end{aligned}
\end{equation}
For \(R\ge0\), every point of \(\mathcal B_j(R)=\{h:d_M(h,\mu_j)\le R\}\) can be written as \(h=\mu_j+Lv\) with \(\|v\|_2\le R\). Let \(v_{jk}=L^{-1}(\mu_j-\mu_k)\). The triangle inequality gives
\begin{equation}
d_M(h,\mu_k)=\|v_{jk}+v\|_2 \le d_M(\mu_k,\mu_j)+R.
\end{equation}
Equality holds at \(v=Rv_{jk}/\|v_{jk}\|_2\) when \(v_{jk}\ne0\), and at any vector of norm \(R\) otherwise. Hence
\begin{equation}
\sup_{h\in\mathcal B_j(R)}d_M(h,\mu_k) =d_M(\mu_k,\mu_j)+R.
\label{eq:supp_enclosing_radius}
\end{equation}
Consequently, the ball \(\{h:d_M(h,\mu_k)\le r\}\) contains \(\mathcal B_j(\sqrt d)\) if and only if \(r\ge d_M(\mu_k,\mu_j)+\sqrt d\). Minimizing this threshold over competing classes yields
\begin{equation}
\begin{aligned}
r_k&=D_k^{\min}+\sqrt d\\
&=\min_{j\ne k}\sup_{h\in\mathcal B_j(\sqrt d)} d_M(h,\mu_k).
\end{aligned}
\end{equation}
Thus \(r_k\) is the smallest Mahalanobis radius about \(\mu_k\) enclosing the reference ellipsoid of a nearest competing class \(j^*\in\arg\min_{j\ne k}d_M(\mu_k,\mu_j)\).

For the second claim, write \(X=\mu_{j^*}+LZ\), where \(Z\sim\mathcal N(0,I_d)\), and let \(D=D_k^{\min}=\|v_{j^*k}\|_2\). Then
\begin{equation}
\begin{aligned}
d_M(X,\mu_k)^2 &=\|v_{j^*k}+Z\|_2^2\\
&=D^2+2v_{j^*k}^{\top}Z+\|Z\|_2^2.
\end{aligned}
\end{equation}
Since \(\mathbb E Z=0\) and \(\mathbb E\|Z\|_2^2=d\), the RMS distance is
\begin{equation}
s_k=\sqrt{\mathbb E\,d_M(X,\mu_k)^2} =\sqrt{D^2+d}.
\end{equation}
Using \(r_k=D+\sqrt d\),
\begin{equation}
\begin{aligned}
r_k^2-s_k^2&=2D\sqrt d\ge0,\\
2s_k^2-r_k^2&=(D-\sqrt d)^2\ge0.
\end{aligned}
\end{equation}
Therefore \(s_k\le r_k\le\sqrt2\,s_k\).

\begin{table*}[t]
\centering
\small
\caption{CRB controls without GRS on CIFAR-10, with the full \proposal{} for reference. 
Values are percentages (mean and standard deviation over three seeds). 
Client selection is measured at broadcast rounds 2000 and 2500 and averaged within each seed. 
Correct and Mislabeled are normalized by the known-class pool, and Unknown by the unknown-class pool. 
PL acc.\ denotes pseudo-label accuracy: the fraction of all selected examples that are correctly labeled known-class examples.}
\label{tab:supp_crb}
\begin{tabular*}{\textwidth}{@{\extracolsep{\fill}}lcccccc@{}}
\toprule
 & \multicolumn{4}{c}{Client selection (\%)} & \multicolumn{2}{c}{Test (\%)} \\
\cmidrule(lr){2-5}\cmidrule(lr){6-7}
Variant & Correct & Mislabeled & Unknown & PL acc. & ACC & AUROC \\
\midrule
Base & 57.0\(\pm\)2.2 & 22.9\(\pm\)1.6 & 70.7\(\pm\)11.3 & 45.0\(\pm\)4.4 & 66.6\(\pm\)2.5 & 75.8\(\pm\)4.3 \\
Post-hoc balancing & 52.4\(\pm\)5.3 & 18.6\(\pm\)3.5 & 56.1\(\pm\)18.4 & 49.4\(\pm\)6.4 & 67.4\(\pm\)2.9 & 78.9\(\pm\)2.4 \\
+CRB & 45.4\(\pm\)9.1 & 16.2\(\pm\)8.2 & 27.2\(\pm\)12.0 & 59.2\(\pm\)8.7 & 68.8\(\pm\)1.7 & 81.9\(\pm\)2.5 \\
+CRB, row-centered & 57.6\(\pm\)4.4 & 21.8\(\pm\)0.8 & 36.9\(\pm\)17.1 & 55.8\(\pm\)6.6 & 68.0\(\pm\)2.9 & 80.4\(\pm\)6.8 \\
\proposal{} & 55.6\(\pm\)2.9 & 17.2\(\pm\)1.0 & 21.9\(\pm\)8.2 & 63.8\(\pm\)5.6 & 69.5\(\pm\)1.7 & 84.0\(\pm\)4.8 \\
\bottomrule
\end{tabular*}
\end{table*}

\subsubsection{Reference Scale and Whitening}
\label{app:reference_scale}

For \(H\sim\mathcal N(\mu_k,\widehat\Sigma)\), \(\mathbb E\,d_M(H,\mu_k)^2=d\). Thus \(\sqrt d\) is an RMS reference radius, not a radius calibrated to a prescribed containment probability.

For \(\|u\|_2=1\) and \(r\ge0\), the reference \(h=\mu_k+rLu\) satisfies
\begin{equation}
d_M(h,\mu_k)=\|L^{-1}(rLu)\|_2=r.
\end{equation}
This identity applies to both positive and negative references in main-paper Eqs.~\eqref{eq:positive_references} and~\eqref{eq:negative_references}. Positive references have their sampled Mahalanobis radii, while negative references lie on the radius-\(r_k\) sphere in whitened coordinates. Both are mapped back to the original feature coordinates for detector supervision.

\subsection{Direct and Indirect Feature Gradients}
\label{app:indirect_gradients}

CRB detaches backbone features and updates only the classifier parameters \(\phi\); GRS detaches reference features and updates only the detector parameters \(\psi\). Neither objective directly updates the backbone \(\theta\), which remains trainable through supervised server and unlabeled client losses.

For \(h=f_\theta(x)\), the cross-entropy feature gradient is
\begin{equation}
\nabla_h\ell_{\rm CE} =J_{g_\phi}(h)^\top \big(\operatorname{softmax}(g_\phi(h))-e_y\big),
\end{equation}
where \(J_{g_\phi}(h)\) is the classifier Jacobian with respect to \(h\), and \(e_y\) is the one-hot target. Updating \(\phi\) can therefore change subsequent feature gradients, as well as future cached targets and confidence masks. Similarly, updating \(\psi\) can change subsequent feature gradients through supervised OVA learning and client SOCR. These paths allow branch-only recalibration to influence later backbone learning indirectly.

%% file: sec/supple/B_CRB.tex
\section{Additional Analysis of CRB}
\label{app:crb_experiments}

Motivated by Proposition~\FBcrbprop{}, we examine whether balancing remains effective after removing common-shift sensitivity and whether post-hoc correction provides comparable benefits.

\noindent
\textbf{Setup.}
We use CIFAR-10 with six known and four unknown classes, 40 labels per known class, and Dirichlet \(\alpha=0.1\). 
All controls disable GRS. Within each seed, labeled splits and client partitions are matched across variants. 
\emph{Row-centered CRB} uses \(\widetilde z\) in both the CRB numerator and denominator (Section~\ref{app:crb_invariance}). 
\emph{Post-hoc balancing} uses the Base server loss and, after each round of fine-tuning, measures \(b\) on the labeled set and broadcasts \(\delta_k=-(b_k-\bar b)\). 
Clients apply this offset only when caching soft targets and confidence masks. 
By Proposition~\FBcrbprop{}(i), this equalizes the competitive responses computed from the calibration logits.

Selection statistics are evaluated on all client pools using the models broadcast at rounds 2000 and 2500, following each variant's target-generation rule, and averaged within each seed. 
Client ground-truth labels are used only for these diagnostics. Test results report known-class accuracy and overall AUROC using the main-paper evaluation protocol. 
The full \proposal{} model is included for reference.

\noindent
\textbf{Results.}
Both CRB forms improve mean pseudo-label accuracy and reduce unknown admission relative to Base (Table~\ref{tab:supp_crb}). 
Raw-logit CRB has the highest mean pseudo-label accuracy and test scores among the controls, but selects fewer correct and mislabeled known examples. 
These known-selection trends differ from Fig.~\ref{tab:ablation} of the main paper, which uses \(\alpha=0.3\) and statistics collected during client training. Row-centered CRB keeps known selection near the Base level, showing that selection benefits persist with a common-shift-invariant objective. Post-hoc balancing also improves selection quality. Compared with it, row-centered CRB retains more correct known examples and admits fewer unknowns, although it also selects more mislabeled known examples. The full \proposal{} has the lowest unknown admission and the highest mean pseudo-label accuracy and test scores.

%% file: sec/supple/C_GRS.tex
\section{Additional Analysis of GRS}
\label{app:grs_experiments}

\begin{table*}[t]
\centering
\small
\caption{Negative-reference radius with CRB enabled (\(\alpha=0.1\)). Without GRS corresponds to the +CRB ablation; the remaining rows use full \proposal{} and vary only the negative-reference radius. Results are single runs with a shared seed per setting. Dataset headings specify (known/unknown classes, labels per known class). ACC and overall AUROC are percentages.}
\label{tab:supp_grs_radius}
\begin{tabular*}{\textwidth}{@{\extracolsep{\fill}}lcccccc@{}}
\toprule
 & \multicolumn{2}{c}{CIFAR-10 (6/4, 40)} & \multicolumn{2}{c}{CIFAR-10 (6/4, 20)} & \multicolumn{2}{c}{CIFAR-100 (80/20, 40)} \\
\cmidrule(lr){2-3}\cmidrule(lr){4-5}\cmidrule(lr){6-7}
Radius & ACC & AUROC & ACC & AUROC & ACC & AUROC \\
\midrule
Without GRS (+CRB) & 70.7 & 84.7 & 59.7 & 78.9 & 56.0 & 65.9 \\
\(0.5\,D_k^{\min}\) & 71.3 & 83.3 & 59.8 & 77.6 & 54.9 & 66.5 \\
\(D_k^{\min}\) & 71.8 & 87.7 & 59.8 & 79.1 & 57.1 & 66.3 \\
\(D_k^{\min}+\sqrt d\) (\textbf{\proposal{}}) & 71.3 & 87.7 & 60.1 & 81.3 & 57.7 & 70.1 \\
\(D^{\max}+\sqrt d\) & 71.2 & 86.6 & 59.1 & 78.9 & 56.8 & 72.9 \\
\bottomrule
\end{tabular*}
\end{table*}

Motivated by Proposition~\FBgrsprop{}, we evaluate sensitivity to the negative-reference radius \(r_k=D_k^{\min}+\sqrt d\).

\noindent
\textbf{Setup.}
We use CIFAR-10 with six known and four unknown classes and 40 or 20 labels per known class, and CIFAR-100 with 80 known and 20 unknown classes, all with Dirichlet \(\alpha=0.1\). All radius variants retain CRB and positive-reference supervision and change only the negative-reference radius. We compare \(r_k\) with \(D_k^{\min}\), \(0.5D_k^{\min}\), and the shared global radius \(D^{\max}+\sqrt d\), where \(D^{\max}=\max_{j\ne\ell}d_M(\mu_j,\mu_\ell)\). The no-GRS reference retains CRB but disables both positive and negative reference supervision. Each entry is a single run, with the seed, labeled split, and client partition matched within each setting. Test results report known-class accuracy and overall AUROC using the main-paper evaluation protocol.

\noindent
\textbf{Results.}
In the reported runs, adding GRS with the proposed radius improves both accuracy and AUROC over +CRB in all three settings (Table~\ref{tab:supp_grs_radius}). Removing the margin matches its AUROC on CIFAR-10 with 40 labels per known class with slightly higher accuracy, but reduces both metrics in the other two settings. Using \(0.5D_k^{\min}\) lowers AUROC relative to the proposed radius in all settings and falls below the no-GRS reference on both CIFAR-10 settings. The larger global radius improves AUROC on CIFAR-100 at the cost of accuracy, while yielding lower accuracy and AUROC on both CIFAR-10 settings. These results support the proposed radius as a geometry-motivated default, with setting-dependent accuracy--AUROC trade-offs.

%% file: sec/supple/D_implementation.tex
\section{\proposal{} Implementation Details}
\label{app:implementation}

This section supplements the method and experimental protocol in Sections~\ref{sec:method} and~\ref{sec:experimental_setup} of the main paper. Exact dataset- and run-specific configurations will be released with the code.

\subsection{Data and Models}
\label{app:data_protocol}

We follow the data protocol in Section~\ref{sec:experimental_setup} of the main paper. Our benchmark datasets and backbone families are also evaluated in FedOpenMatch~\cite{liu2026fedopenmatch}. CIFAR-100 known classes belong to superclasses 0--10 (55/45) or 0--15 (80/20). SVHN uses only the standard training split. The full unlabeled pool, containing both known and unknown classes, is partitioned using class-wise Dirichlet allocation followed by 100 iterations of row and column normalization. IID partitions use \(\alpha=10^5\).

Both branch MLPs have two hidden layers with biases and ReLU activations, followed by bias-free outputs. Classifier hidden widths are 128, 1024, and 512 on CIFAR-10, CIFAR-100, and SVHN, respectively; the detector width is 1024. The SVHN ResNet-18 uses a \(3\times3\) stride-1 stem without max pooling. For \proposal{} and all baselines except SemiFL, evaluation uses an exponential moving average (EMA) of model parameters and batch normalization (BN) buffers with decay 0.9, updated once per communication round after server fine-tuning. This smooths round-to-round model fluctuations. SemiFL is the exception to the main-paper statement that all methods are evaluated with EMA models: it is evaluated with its final global model, as in its official implementation.

\subsection{Training Procedure}
\label{app:training_algorithm}

Algorithm~\ref{alg:supp_fedbare} summarizes training. Let \(\omega=(\theta,\phi,\psi)\) collect the backbone, classifier, and detector parameters, and let \(\bar\omega\) denote their EMA. The participating clients at round \(t\) form \(\mathcal C_t\).

For data augmentation, \(x^{w_1}\) uses random horizontal flipping, while \(x^{w_2}\) additionally applies random cropping with four-pixel reflection padding. The strong view \(x^s\) adds RandAugmentMC (\(n=2,m=10\)), including Cutout. Before local training, each client caches soft classifier targets \(p_i\) and confidence masks \(m_i\) from the broadcast model on \(x^{w_1}\), as defined in main-paper Eq.~\eqref{eq:client_objective}. These remain fixed during local training, while input views are resampled at each step.

For GRS, \(S\) is the shared within-class covariance, \(\widehat\Sigma\) is its OAS-regularized version in main-paper Eq.~\eqref{eq:feature_geometry}, and \(\hat\rho\) is the pooled 95th percentile of labeled features' own-class Mahalanobis distances. The sets \(\mathcal R_k^+\) and \(\mathcal R_k^-\) provide positive and negative feature references for detector head \(k\). Server losses use two labeled weak views; two strong views also enter the joint forward pass and affect BN statistics without a direct loss. Geometry refitting and the activation of the auxiliary losses follow the
schedule in Section~\ref{app:crb_grs_details}.

\begin{figure*}[t]
\centering
\begin{minipage}{\textwidth}
\small
\hrule\vspace{4pt}
\refstepcounter{FBalgorithm}\label{alg:supp_fedbare}
\textbf{Algorithm~\theFBalgorithm. \proposal{} training}
\par\vspace{4pt}\hrule\vspace{4pt}
\noindent\textbf{Input:} labeled server set \(\mathcal D_s\) with \(K\) known classes, unlabeled client pools \(\{\mathcal D_c\}\), rounds \(R\), GRS start round \(t_{\mathrm{GRS}}\), and GRS start epoch \(e_{\mathrm{GRS}}\).\par
\noindent\textbf{Output:} EMA evaluation model \(\bar\omega^R\).\par
\begin{list}{\arabic{FBalgline}:}{\usecounter{FBalgline}\setlength{\leftmargin}{2.2em}\setlength{\labelwidth}{1.7em}\setlength{\labelsep}{0.5em}\setlength{\itemsep}{0pt}\setlength{\parsep}{0pt}\setlength{\topsep}{3pt}}
\item Obtain \(\omega^0\) and its EMA \(\bar\omega^0\) by 500 supervised initialization epochs on \(\mathcal D_s\) with \(\mathcal L_{\rm CE}+\lambda_{\rm OVA}\mathcal L_{\rm OVA}\), updating the EMA after each epoch.
\item Refresh the BN running statistics of \(\omega^0\) with a gradient-free pass over \(\mathcal D_s\).
\item \textbf{for} round \(t=1,\ldots,R\) \textbf{do}
\item \hspace*{1em}Sample ten clients \(\mathcal C_t\) and broadcast \(\omega^{t-1}\).
\item \hspace*{1em}\textbf{for each} \(c\in\mathcal C_t\) \textbf{in parallel do}
\item \hspace*{2em}Cache \(\{p_i,m_i\}_{i\in\mathcal D_c}\) with \(\omega^{t-1}\) in evaluation mode.
\item \hspace*{2em}Obtain \(\omega_c\) from \(\omega^{t-1}\) by five local epochs of main-paper Eq.~\eqref{eq:client_objective}.
\item \hspace*{1em}\textbf{end for}
\item \hspace*{1em}Aggregate \(\displaystyle \omega_s\leftarrow\frac{\sum_{c\in\mathcal C_t}|\mathcal D_c|\,\omega_c}{\sum_{c\in\mathcal C_t}|\mathcal D_c|}\).
\item \hspace*{1em}\textbf{for} server epoch \(e=1,2,3\) \textbf{do}
\item \hspace*{2em}\textbf{if} \(t\ge t_{\mathrm{GRS}}\) and \(e\ge e_{\mathrm{GRS}}\), estimate class means \(\{\mu_k\}_{k=1}^K\) and covariance \(S\) from current labeled server features, then compute \(\widehat\Sigma\) and \(\hat\rho\).
\item \hspace*{2em}\textbf{for each} labeled minibatch \textbf{do}
\item \hspace*{3em}\textbf{if} \(t\ge t_{\mathrm{GRS}}\) and \(e\ge e_{\mathrm{GRS}}\), sample \(\{\mathcal R_k^+,\mathcal R_k^-\}_{k=1}^K\) using main-paper Eqs.~\eqref{eq:positive_references}--\eqref{eq:negative_references}.
\item \hspace*{3em}Update \(\omega_s\) with main-paper Eq.~\eqref{eq:server_objective}, setting inactive auxiliary terms to zero.
\item \hspace*{2em}\textbf{end for}
\item \hspace*{1em}\textbf{end for}
\item \hspace*{1em}Update \(\bar\omega^t\leftarrow0.9\bar\omega^{t-1}+0.1\omega_s\).
\item \hspace*{1em}Update the BN running statistics of \(\omega_s\) with a gradient-free pass over \(\mathcal D_s\), and set \(\omega^t\leftarrow\omega_s\).
\item \textbf{end for}
\end{list}
\vspace{3pt}\hrule
\end{minipage}
\end{figure*}

\subsection{CRB and GRS Details}
\label{app:crb_grs_details}

CRB uses population variances with denominator stabilizer \(\epsilon=10^{-6}\) in main-paper Eq.~\eqref{eq:crb_loss}. GRS geometry is fitted using evaluation-mode features of \(x^{w_2}\). The covariance is normalized by the number of fitted samples minus the number of represented classes. In main-paper Eq.~\eqref{eq:feature_geometry}, we set the diagonal jitter to \(\eta=10^{-5}\). The radius \(\hat\rho\) is the 95th percentile of \(d_M(h_i,\mu_{y_i})\), pooled over labeled examples.

Within each class, positive directions follow labeled samples ranked by decreasing backbone feature norm, cycling when fewer labels than candidates are available.

\noindent
\textbf{Hyperparameters and schedule.}
We fix $\lambda_{\mathrm{OVA}}=\lambda_{\mathrm{SOCR}}=1$ in all experiments.
We set $(\lambda_{\mathrm{CRB}},\lambda_{\mathrm{GRS}})$ to $(1.0,\,0.01)$ on CIFAR-10
with 40 labels per class and on SVHN, $(0.5,\,0.05)$ on CIFAR-10 with 20 labels per
class, and $(0.03,\,0.001)$ on CIFAR-100.
These weights were chosen so that the weighted auxiliary terms stay on the same
order as $\mathcal{L}_{\mathrm{CE}}$ and $\mathcal{L}_{\mathrm{OVA}}$; in particular,
$\mathcal{L}_{\mathrm{GRS}}$ sums over the $K$ heads
(main-paper Eq.~\eqref{eq:geometric_loss}), so $\lambda_{\mathrm{GRS}}$ is smaller on CIFAR-100.
CRB is applied in every server epoch from the first round.
GRS is enabled from round $t_{\mathrm{GRS}}=50$, $100$, and $300$ in these three groups, respectively, and from server epoch $e_{\mathrm{GRS}}=2$ of each round in all experiments, with the geometry refitted before each GRS epoch.
Each server step draws 128 negative references per head and one positive
candidate per labeled feature of the class.
Clients use temperature $T=1$ and threshold $\tau=0.95$ on CIFAR-10 and SVHN, and
$T=0.5$ and $\tau=0.99$ on CIFAR-100.

%% file: sec/supple/E_baseline.tex
\section{Baseline Implementations and Verification}
\label{app:baseline_implementation}
\label{sec:supp_baseline_verification}

We distinguish existing federated baselines, SemiFL and FedOpenMatch, from our federated adaptations of FixMatch, SSB, SCOMatch, and ProSub. For each baseline, we describe its training objectives, prediction heads, and how its objectives are applied during server and client training. The EMA evaluation described in Section~\ref{app:data_protocol} applies to main-table comparisons, except for SemiFL; original-protocol reproductions use the evaluation models specified below.

\subsection{Existing Federated Baselines}
\label{app:existing_federated_baselines}
\label{app:baseline_reproduction}
\label{app:baseline_failures}

\begin{figure*}[t]
\centering
\includegraphics[width=\textwidth]{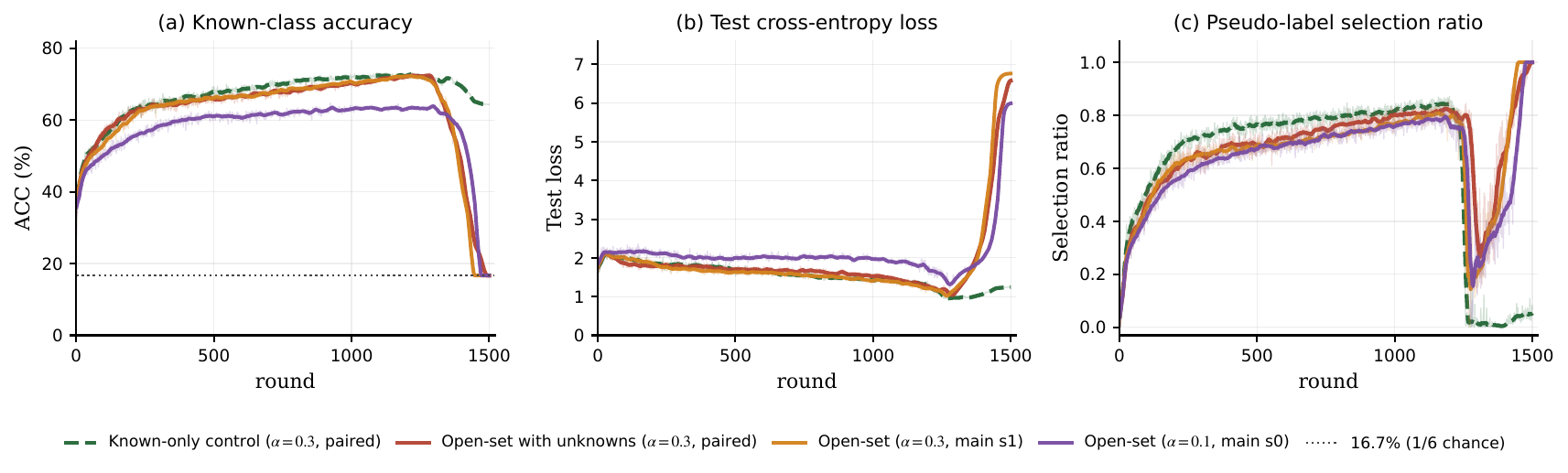}
\caption{SemiFL on CIFAR-10 with 40 labels per known class over 1,500 rounds. The matched known-only and open-set runs use an internal \(\alpha=0.3\) partition; additional open-set curves use main-table seed~1 at \(\alpha=0.3\) and seed~0 at \(\alpha=0.1\). Panels show (a) known-class accuracy, (b) test cross-entropy loss, and (c) the confidence-based pseudo-label selection ratio. Thin lines show per-round values and thick lines show 21-round moving averages.}
\label{fig:supp_semifl_closed_vs_open}
\end{figure*}

\begin{figure*}[t]
\centering
\includegraphics[width=\textwidth]{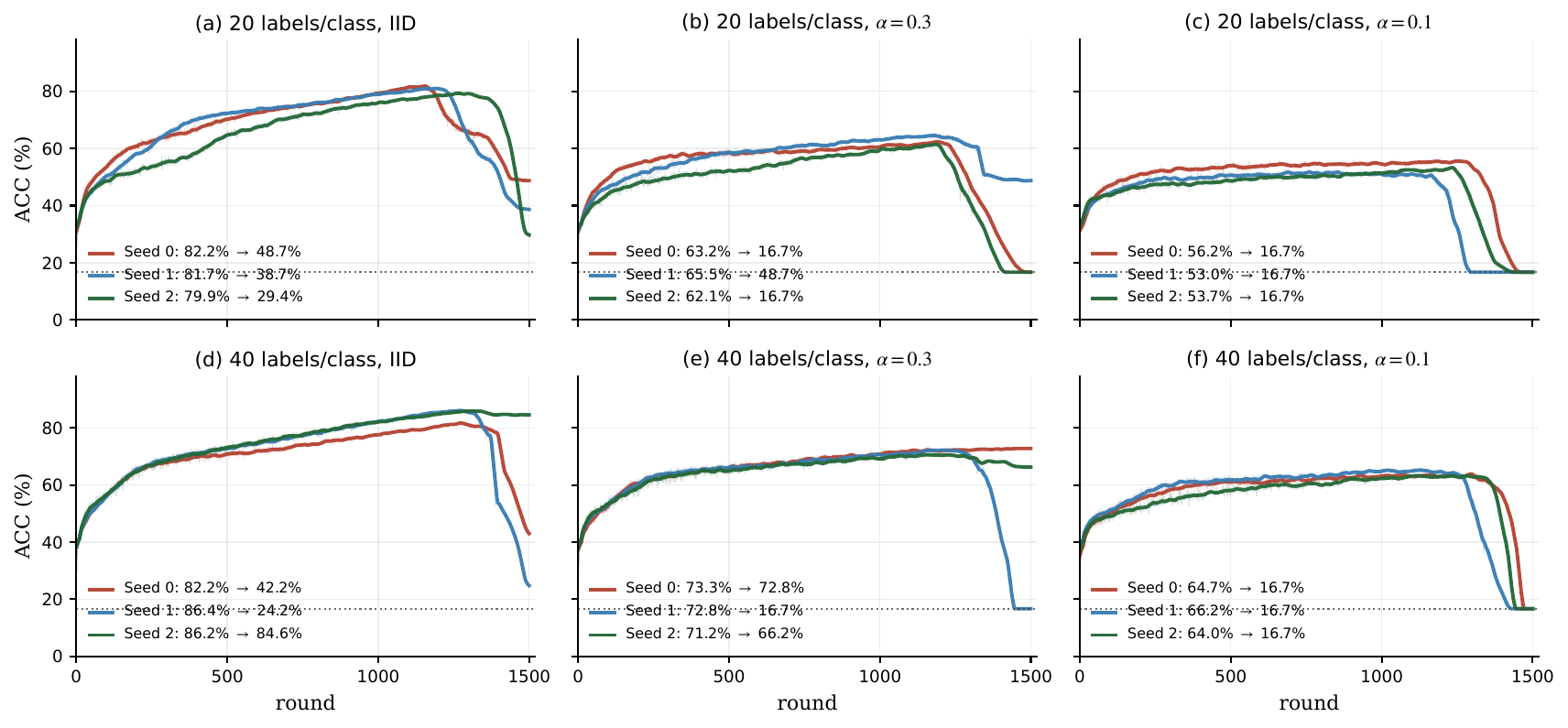}
\caption{Known-class accuracy trajectories of all CIFAR-10 SemiFL runs corresponding to main-paper Table~\ref{tab:main_results}. Each setting contains three seeds evaluated over 1,500 rounds. Thin lines show per-round values and thick lines show 21-round moving averages. The dotted line marks chance-level accuracy for six known classes.}
\label{fig:supp_semifl_openset_curves}
\end{figure*}

\subsubsection{SemiFL}
\label{app:baseline_semifl}

\noindent
\textbf{Implementation.}
We use the official SemiFL~\cite{diao2022semifl} implementation, matching the backbone, data splits, label budgets, and client participation of \proposal{}. Specifically, we use WRN-28-2 on CIFAR-10/100 and ResNet-18 on SVHN, with 100 clients and 10\% random participation per round. Each round runs five local client epochs using FixMatch-MixUp (\(\tau=0.95\), MixUp \(\alpha_{\rm mix}=0.75\)) with static batch normalization (sBN) and batch size 10 (32 on SVHN), followed by server-side aggregation with global SGD momentum (0.5) and five server fine-tuning epochs on the labeled set. Local optimization uses SGD with initial learning rate 0.03, Nesterov momentum 0.9, weight decay \(5\times 10^{-4}\), and cosine learning-rate decay. The CIFAR runs use 1,500 communication rounds instead of the default 2,500.
In preliminary runs, SemiFL's known-class accuracy plateaued within 1,500 rounds and longer training was less stable, so we adopt the shorter schedule.

OOD detection is evaluated using \(1-\mathrm{MSP}\).

\noindent
\textbf{Closed-set verification.}
We compare a matched pair of 1,500-round runs on CIFAR-10 with six known classes, 40 labels per class, and Dirichlet \(\alpha=0.3\), using an internal SemiFL partition separately from the main-table benchmark splits. The known-only control removes the four unknown classes from the client pools while retaining the six-class classifier and training schedule. Collapse occurs in the open-set run but not in the known-only control (Figure~\ref{fig:supp_semifl_closed_vs_open}; top block of Table~\ref{tab:supp_semifl_summary}).

\noindent
\textbf{Observed behavior under our protocol.}
Figure~\ref{fig:supp_semifl_openset_curves} and the bottom block of Table~\ref{tab:supp_semifl_summary} summarize the CIFAR-10 runs reported in the main paper. All runs initially learn, but late-stage degradation varies across settings and seeds; all \(\alpha=0.1\) runs finish at chance-level known-class accuracy. The completely collapsed runs show saturation of pseudo-label selection and increased test cross-entropy (Figure~\ref{fig:supp_semifl_closed_vs_open}). No collapse is observed in the reported CIFAR-100 runs during the same 1,500-round schedule.

\begin{table*}[t]
\centering
\small
\setlength{\tabcolsep}{3pt}
\caption{SemiFL 1,500-round diagnostic summary on CIFAR-10. Top: matched 1,500-round known-only control versus open-set run on SemiFL's internal \(\alpha=0.3\) partition (40 labels per class, separate from the main-table benchmark splits). Bottom: peak versus final known-class accuracy (mean \(\pm\) standard deviation over three seeds) for all 18 main-table CIFAR-10 open-set runs (1,500 rounds), together with collapse counts (final accuracy \(\le 18\%\)), mean final pseudo-label selection ratio (Sel.\ ratio), and mean final test loss.}
\label{tab:supp_semifl_summary}
\begin{tabular*}{\textwidth}{@{\extracolsep{\fill}}lcccccc@{}}
\toprule
Setting (CIFAR-10, WRN-28-2) & Rounds & Peak Acc. (\%) & Final Acc. (\%) & Collapsed & Sel.\ ratio & Final Loss \\
\midrule
\multicolumn{7}{l}{\textit{Matched 1,500-round control pair (40 labels per class, \(\alpha=0.3\), paired internal partition)}} \\
Known-only control (6 classes, no unknowns) & 1,500 & 73.3 & 64.4 & 0/1 & 0.07 & 1.25 \\
Open-set with unknowns (6/4 split, paired) & 1,500 & 73.1 & 16.7 & 1/1 & 1.00 & 6.59 \\
\midrule
\multicolumn{7}{l}{\textit{Main-table CIFAR-10 open-set runs (6/4 split, three seeds each; matching Table~\ref{tab:main_results} of the main paper)}} \\
20 labels/class, IID & 1,500 & \(81.2\pm1.2\) & \(38.9\pm9.7\) & 0/3 & 0.78 & 3.72 \\
20 labels/class, \(\alpha=0.3\) & 1,500 & \(63.6\pm1.7\) & \(27.4\pm18.5\) & 2/3 & 0.79 & 5.64 \\
20 labels/class, \(\alpha=0.1\) & 1,500 & \(54.3\pm1.7\) & \(16.7\pm0.0\) & 3/3 & 1.00 & 7.35 \\
40 labels/class, IID & 1,500 & \(85.0\pm2.4\) & \(50.3\pm31.0\) & 0/3 & 0.60 & 2.81 \\
40 labels/class, \(\alpha=0.3\) & 1,500 & \(72.4\pm1.1\) & \(51.9\pm30.7\) & 1/3 & 0.34 & 2.96 \\
40 labels/class, \(\alpha=0.1\) & 1,500 & \(65.0\pm1.1\) & \(16.7\pm0.0\) & 3/3 & 1.00 & 6.66 \\
\bottomrule
\end{tabular*}
\end{table*}

\begin{table}[t]
\centering
\caption{Reproduction of FedOpenMatch with the data split and data setting of its paper (mean $\pm$ standard error over seeds 0, 10, and 100, as in the paper).
Server-side BN recalibration uses labeled data only.
$\alpha$ is the Dirichlet parameter of the client partition.
\emph{Paper} denotes the values reported in the FedOpenMatch paper and \emph{Repro.}\ our reproduction under the same setting.}
\label{tab:supp_fedopenmatch_repro}
\setlength{\tabcolsep}{4pt}
\newcommand{\fomms}[2]{\begin{tabular}[c]{@{}c@{}}#1\\[-3pt]{\scriptsize\(\pm\)#2}\end{tabular}}
\resizebox{\columnwidth}{!}{%
\begin{tabular}{llcccc}
\toprule
Dataset & & \multicolumn{2}{c}{Closed-set acc. (\%)} & \multicolumn{2}{c}{Open-set acc. (\%)} \\
\cmidrule(lr){3-4}\cmidrule(lr){5-6}
(known/labels) & \(\alpha\) & Paper & Repro. & Paper & Repro. \\
\midrule
CIFAR-10  & 0.1 & \fomms{65.19}{2.83} & \fomms{69.64}{0.58} & \fomms{55.27}{2.21} & \fomms{59.67}{0.52} \\
(6/40)    & 0.3 & \fomms{73.32}{1.12} & \fomms{76.74}{0.41} & \fomms{62.26}{0.51} & \fomms{65.68}{0.35} \\
\midrule
CIFAR-100 & 0.1 & \fomms{37.11}{0.18} & \fomms{38.44}{0.41} & \fomms{36.65}{2.43} & \fomms{36.21}{0.84} \\
(80/10)   & 0.3 & \fomms{39.45}{1.04} & \fomms{40.78}{0.77} & \fomms{38.97}{1.03} & \fomms{39.38}{1.69} \\
\midrule
SVHN      & 0.1 & \fomms{82.63}{1.50} & \fomms{90.86}{1.20} & \fomms{70.57}{0.93} & \fomms{78.10}{1.05} \\
(6/40)    & 0.3 & \fomms{84.86}{1.57} & \fomms{94.17}{0.50} & \fomms{71.87}{0.83} & \fomms{80.74}{0.36} \\
\bottomrule
\end{tabular}}
\end{table}

\subsubsection{FedOpenMatch}
\label{app:baseline_fedopenmatch}
\label{app:ova_scoring_audit}

\noindent
\textbf{Implementation.}
FedOpenMatch~\cite{liu2026fedopenmatch} is the direct FOSSL comparison and uses a classifier with a class-wise OVA detector. We use the implementation provided in the supplementary material of the paper and port its training objectives to the Flower framework~\cite{beutel2020flower}. The server trains with classifier cross-entropy and an OVA loss using hard-negative sampling and logit adjustment. OVA objectives use detached backbone features. At the start of each round, clients generate pseudo-labels using the global model and keep them fixed during local training. Client training combines confident-inlier cross-entropy, OVA pseudo-label supervision, and logit consistency regularization. A sample is selected as a confident inlier when its classifier confidence is at least 0.95 and the OVA inlier probability of its predicted class is at least 0.5. OVA pseudo-labels use a positive threshold of 0.99 and a negative threshold of 0.01 (0.1 on CIFAR-100, as in the official scripts). The logit consistency regularization, confident inlier, and OVA loss coefficients are all 1.0. We modify the treatment of batch normalization statistics. The original implementation also uses client unlabeled data to recompute batch-normalization statistics. We use only labeled server data because client data are not accessible to the server in the federated setting. Federated training settings such as client sampling and FedAvg aggregation match those of \proposal{}. The outlier score is the OVA outlier probability of the predicted class.

\noindent\textbf{Reproduction and verification.}
To verify our FedOpenMatch implementation, we reproduce the results reported in the original paper using the same Flower implementation as in our main experiments. The protocol used in our main experiments differs from the original FedOpenMatch protocol in several respects, including client data partitioning. We sample Dirichlet proportions separately for each known and unknown class and balance client sizes. In contrast, the official implementation groups all unknown classes together and allows unequal client sizes. Some of our label budgets and known/unknown class splits also differ from those reported in the FedOpenMatch paper. For reproduction, we select one combination of known/unknown class split and label budget from the original paper for each dataset and evaluate both Dirichlet settings ($\alpha \in \{0.1, 0.3\}$). We use the official data partitioning code and training configurations, including the backbone, optimizer, number of clients, and learning rate. The server aggregation follows the official implementation, using equal-weight averaging and server momentum. We retain the modification described above and recompute batch-normalization statistics using only labeled server data. Following the original FedOpenMatch evaluation protocol, we report closed-set and open-set accuracy. For open-set accuracy, all unknown test classes are merged into a single class. A sample is classified as unknown if the OVA outlier probability for its predicted class exceeds 0.5. Balanced accuracy is then computed over the \(K{+}1\) classes. We evaluate the final global model without EMA, as in the official implementation.

Our reproduction achieves higher closed-set accuracy than reported in the original paper across the tested settings. Open-set accuracy is also comparable to or higher than reported values (Table~\ref{tab:supp_fedopenmatch_repro}). These results support the validity of our FedOpenMatch implementation under the tested reproduction settings. The performance gap between the reproduction and our main experiments, particularly on SVHN, likely reflects differences in data partitioning, class splits, label budgets, model backbones, and training configurations.

\begin{figure*}[t]
\centering
\includegraphics[width=\textwidth]{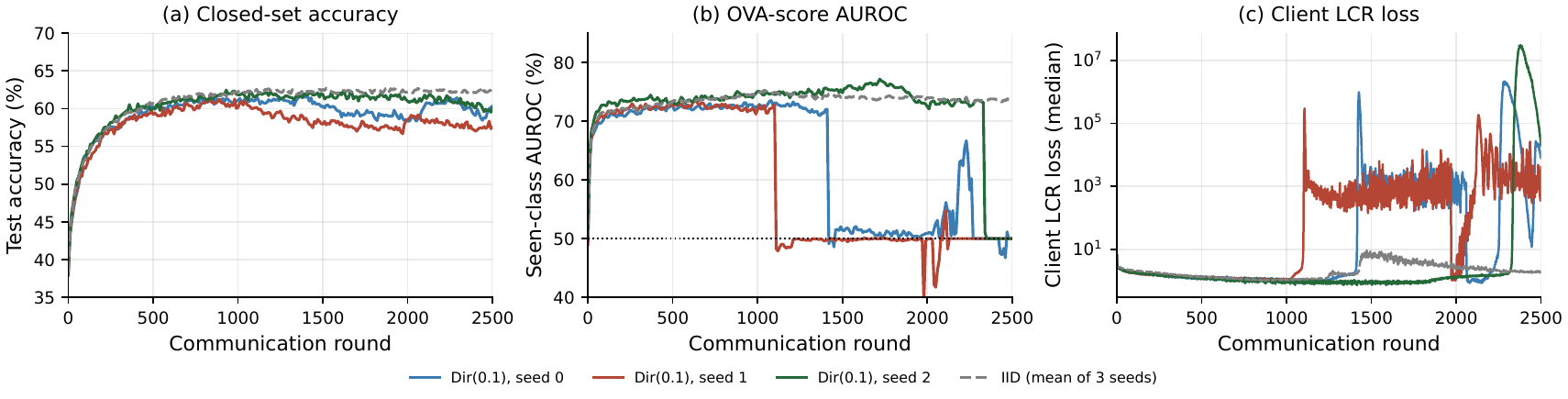}
\caption{Training curves of FedOpenMatch on CIFAR-100 with 55 known classes under Dir(0.1). (a) Closed-set accuracy and (b) seen-OOD AUROC of the OVA outlier score, evaluated every 10 rounds. (c) Logit-consistency loss of the participating clients (median per round, log scale). The dashed gray line is the mean over three seeds under IID partitions, and the dotted line in (b) marks chance-level AUROC.}
\label{fig:supp_fom_collapse}
\end{figure*}

\noindent
\textbf{Observed behavior under our protocol.}
Under the CIFAR-100 55/45 split with Dirichlet \(\alpha=0.1\), the reported probability-based AUROC falls to chance level in every seed (Table~\ref{tab:main_results}). The decline follows an initial improvement and coincides with rapid growth in OVA logit magnitudes and saturation of the outlier probabilities, leaving all test samples with the same score (Figure~\ref{fig:supp_fom_collapse}). The probability-based score remains saturated through the final round, while known-class accuracy remains above chance. A similar failure occurs in one seed at \(\alpha=0.3\), but is not observed in the IID runs.

\subsection{Federated Adaptations of SSL and OSSL Methods}
\label{app:baseline_adaptations}

FedFixMatch adapts a centralized SSL method, while FedSSB, FedSCOMatch, and FedProSub adapt centralized OSSL methods. For each adaptation, we place labeled-data objectives at the server and unlabeled-data objectives at clients, with the method-specific state handling described below. The source papers' centralized results are distinct from the federated results reported in our main tables.

\subsubsection{FedFixMatch}
\label{app:baseline_fedfixmatch}

\noindent
\textbf{Implementation.}
FedFixMatch uses supervised cross-entropy at the server. At the beginning of each round, clients use the broadcast model to cache pseudo-labels from weakly augmented inputs, which remain fixed during local training. Clients train the classifier on strong views using the cached targets and confidence-based selection, without SOCR.

\begin{figure}[t]
\centering
\includegraphics[width=.8\linewidth]{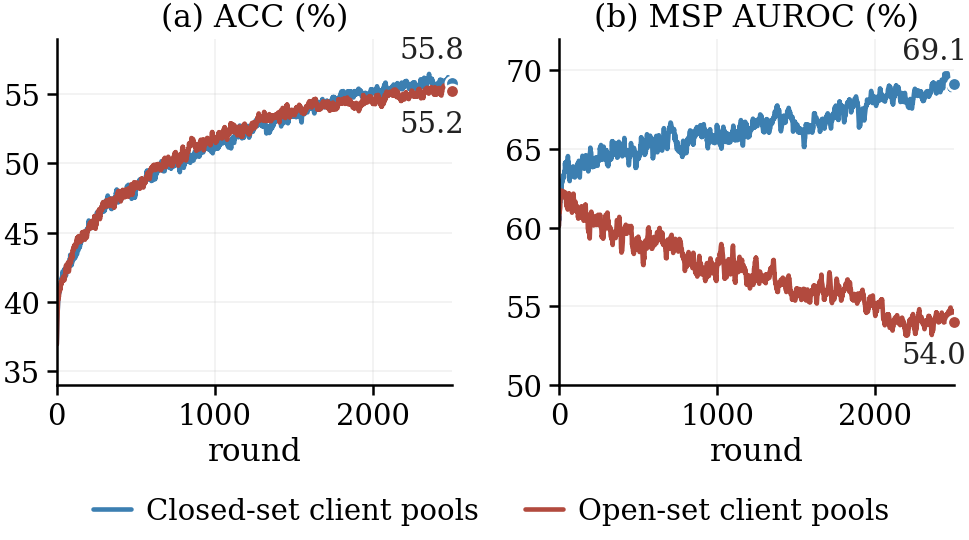}
\caption{FedFixMatch on CIFAR-100 with 80 known classes (\(\alpha=0.3\)) with closed-set and open-set client pools. Closed-set pools contain known-class images only; all other settings, the seed, and the labeled split are shared. Both runs are evaluated on the same test set, and MSP AUROC is computed between known- and unknown-class test images. The numbers mark the values at the final round.}
\label{fig:supp_fedfixmatch_closed}
\end{figure}

\noindent
\textbf{Closed-set verification.}
To verify the implementation, we train FedFixMatch on CIFAR-100 with 80 known classes and Dirichlet \(\alpha=0.3\) after removing all unknown-class images from the client pools, keeping every other setting unchanged. Figure~\ref{fig:supp_fedfixmatch_closed} compares this closed-set run with the open-set run. Accuracy follows a similar curve in both runs. With closed-set pools, MSP AUROC increases throughout training. With open-set pools, it declines while accuracy continues to increase. In this matched comparison, MSP AUROC declines with open-set client pools but not in the known-only control.

\subsubsection{FedSSB}
\label{app:baseline_fedssb}

\noindent
\textbf{Implementation.}
FedSSB uses the same supervised classifier cross-entropy as \proposal{} at the server. Its supervised OVA loss follows the original SSB implementation, treating the target-class head as positive and all remaining heads as negatives. A hard-negative variant produces comparable performance. At each round's start, a frozen, evaluation-mode copy of the broadcast model generates cached pseudo-labels and head-wise pseudo-negative masks from crop-and-flip views. Classifier consistency uses cached hard pseudo-labels on strong views, with the temperature and confidence threshold matched to \proposal{}. Pseudo-negative masks use an OVA inlier-probability threshold of 0.01. These targets and masks remain fixed during local training. The detector uses entropy minimization and SOCR on two weak views (crop-and-flip and flip-only), and pseudo-negative supervision on the strong view. The entropy and pseudo-negative coefficients are 0.1 and 1.0; the SOCR coefficient is 1.0. The overall detector-loss coefficient is 1.0. All other federated training settings, including client sampling and FedAvg aggregation, match those of \proposal{}.

\subsubsection{FedSCOMatch}
\label{app:baseline_fedscomatch}

\noindent
\textbf{Implementation.}
We adapt the official SCOMatch~\citep{wang2024scomatch} implementation, which uses a (\(K{+}1\))-way classifier whose last class represents outliers. The server trains the model with cross-entropy on labeled known-class samples. At the start of each round, clients generate open-set pseudo-labels over all \(K{+}1\) classes and closed-set pseudo-labels over the \(K\) known classes using the broadcast global model. Both remain fixed during local training. Closed-set pseudo-labels are used only for samples whose open-set prediction belongs to a known class. Both self-training objectives use a confidence threshold of 0.95. For the outlier class, we use the same fixed threshold instead of the adaptive threshold of SCOMatch.
Each client maintains a local queue of samples with the lowest maximum known-class probability. The queue persists across rounds and is never shared with the server. Queued samples are trained toward the outlier class using the OOD supervision loss. The OOD supervision loss coefficient is 1.0, as are the coefficients for open-set and closed-set self-training. The outlier score is the probability of the (\(K{+}1\))-th class. Federated training settings such as client sampling and FedAvg aggregation match those of \proposal{}.

\subsubsection{FedProSub}
\label{app:baseline_fedprosub}

\noindent
\textbf{Implementation.}
We adapt the official ProSub~\citep{wallin2024prosub} implementation, which measures the alignment between a feature and the subspace spanned by inlier class prototypes. ProSub models the resulting subspace scores of inliers and outliers using separate Beta distributions, initialized as $\mathrm{Beta}(10,2)$ and $\mathrm{Beta}(2,10)$, respectively, with an inlier prior of 0.5. For OOD detection, we use one minus the subspace score. The server trains the model on labeled data with cross-entropy and updates class prototypes using an EMA of labeled feature means with coefficient 0.99. It also estimates the parameters of the inlier Beta distribution from the subspace scores of labeled features. At each round, the server broadcasts the prototypes and distribution parameters to clients. Clients sample inlier masks from posterior inlier probabilities computed using the two Beta distributions and generate pseudo-labels with a confidence threshold of 0.95. The subspace loss uses the sampled ID/OOD masks to increase scores for inliers and decrease subspace scores for outliers. The pseudo-labeling loss applies only to confident sampled inliers. Following SemiFL~\cite{diao2022semifl}, both pseudo-labels and masks remain fixed during local training. Self-supervision uses all unlabeled samples, regardless of their sampled ID/OOD assignments. Clients optimize the self-supervision, pseudo-labeling, and subspace losses with coefficients 5.0, 1.0, and 1.0, respectively. As in ProSub, the latter two losses are disabled during an initial training phase. During local training, each client estimates the parameters of both Beta distributions from the subspace scores of its unlabeled features, weighted by their posterior inlier and outlier probabilities. After each local epoch, the client updates its distribution parameters using an EMA of these estimates with coefficient 0.99. Once local training is complete, the server averages the clients' distribution parameters and incorporates the result into its global distribution parameters using an EMA with coefficient 0.99. Federated training settings such as client sampling and FedAvg aggregation match those of \proposal{}.

%% file: sec/supple/F_expanded_eval.tex
\section{Expanded Evaluation}
\label{app:additional_results}

This section complements the main results with repeated component ablations and separate seen and unseen OOD evaluation on CIFAR.
We also examine sensitivity to loss weights and performance with fewer labeled server examples.

\begin{table}[ht]
\centering
\small
\setlength{\tabcolsep}{3pt}
\caption{Repeated component ablations under Dirichlet \(\alpha=0.1\). Each cell reports the mean (top) and standard deviation (bottom) over three seeds. ACC and overall AUROC are percentages. Class splits specify known/unknown counts. Best and second-best means are bold and underlined, respectively.}
\label{tab:supp_repeat_ablation}
\providecommand{\ablres}[2]{\shortstack{#1\\[-1pt]{\scriptsize\(\pm\)#2}}}
\begin{tabular*}{\columnwidth}{@{\extracolsep{\fill}}lcccc@{}}
\toprule
\multicolumn{5}{l}{\textit{CIFAR-10 (6/4)}} \\
 & \multicolumn{2}{c}{40 labels/class} & \multicolumn{2}{c}{20 labels/class} \\
\cmidrule(lr){2-3}\cmidrule(lr){4-5}
Variant & ACC & AUROC & ACC & AUROC \\
\midrule
Base & \ablres{66.6}{2.5} & \ablres{75.8}{4.3} & \ablres{53.8}{3.5} & \ablres{73.4}{4.6} \\
+CRB & \ablres{68.8}{1.7} & \ablres{81.9}{2.5} & \ablres{\underline{58.4}}{1.1} & \ablres{74.7}{3.7} \\
+GRS & \ablres{\underline{69.4}}{1.7} & \ablres{\underline{82.7}}{4.2} & \ablres{57.9}{1.6} & \ablres{\underline{77.0}}{3.5} \\
\proposal{} & \ablres{\textbf{69.5}}{1.7} & \ablres{\textbf{84.0}}{4.8} & \ablres{\textbf{60.2}}{1.2} & \ablres{\textbf{78.1}}{4.1} \\
\midrule
\multicolumn{5}{l}{\textit{CIFAR-100 (40 labels/class)}} \\
 & \multicolumn{2}{c}{55/45} & \multicolumn{2}{c}{80/20} \\
\cmidrule(lr){2-3}\cmidrule(lr){4-5}
Variant & ACC & AUROC & ACC & AUROC \\
\midrule
Base & \ablres{63.2}{0.4} & \ablres{73.6}{1.8} & \ablres{54.8}{1.0} & \ablres{69.2}{2.3} \\
+CRB & \ablres{\underline{63.6}}{1.2} & \ablres{73.8}{2.8} & \ablres{55.6}{0.6} & \ablres{69.6}{3.3} \\
+GRS & \ablres{63.2}{1.0} & \ablres{\textbf{76.2}}{2.9} & \ablres{\underline{56.0}}{0.9} & \ablres{\underline{70.7}}{1.8} \\
\proposal{} & \ablres{\textbf{64.0}}{0.4} & \ablres{\underline{76.0}}{2.0} & \ablres{\textbf{56.9}}{1.1} & \ablres{\textbf{70.9}}{1.3} \\
\bottomrule
\end{tabular*}
\end{table}

\subsection{Repeated Component Ablations}

Table~\ref{tab:supp_repeat_ablation} extends the single-run component ablation in main-paper Table~\ref{tab:ablation} to repeated evaluation across seeds.
The repeated-run means preserve the main qualitative pattern: the combined method achieves the highest accuracy across settings and the highest AUROC except on the CIFAR-100 55/45 split, where GRS alone leads.
Relative to Base, \proposal{} improves both mean accuracy and mean AUROC in every setting. Each objective alone also raises mean AUROC in every setting, although the CRB gains are marginal on CIFAR-100.

Base can also be compared with FedSSB at \(\alpha=0.1\) in main-paper Table~\ref{tab:main_results}. Unlike FedSSB, Base applies no entropy minimization or pseudo-negative mining at clients, so the detector receives no client-generated targets. Its means are slightly higher on CIFAR-10 and close on CIFAR-100, with all differences smaller than the across-seed standard deviation of Base. Omitting the client-side detector objectives thus does not reduce performance in these settings, and the gains of CRB and GRS are measured over a base that is already comparable to the strongest detection baseline.

\subsection{Sensitivity to Loss Weights}
\label{app:weight_sensitivity}

We vary one auxiliary-loss weight at a time, holding the remaining training settings fixed.
The CRB sweep reports three-seed results on CIFAR-10, while the GRS sweep uses single-seed runs on CIFAR-10 and CIFAR-100.
Within each sweep, variants share the labeled split, client partition, and initialization seed for each evaluated seed.
Tables~\ref{tab:supp_weight_crb} and~\ref{tab:supp_weight_grs} specify the settings and tested weights.

With GRS fixed at its default setting, all tested positive CRB weights yield higher mean AUROC than disabling CRB, while mean accuracy remains similar across the tested weights (Table~\ref{tab:supp_weight_crb}).
In the GRS sweep, several tested weights improve both accuracy and AUROC over disabling GRS on both datasets (Table~\ref{tab:supp_weight_grs}).
These gains persist across a range of weights, although larger GRS weights can reduce AUROC.

\begin{table}[ht]
\centering
\small
\setlength{\tabcolsep}{5pt}
\caption{Sensitivity to the CRB weight on CIFAR-10 with 40 labels per known class and Dirichlet \(\alpha=0.1\).
Entries report mean \(\pm\) standard deviation over three seeds.
GRS remains enabled at its default setting.
ACC and overall AUROC are percentages. 
The zero-weight row corresponds to +GRS.}
\label{tab:supp_weight_crb}
\begin{tabular*}{\columnwidth}{@{\extracolsep{\fill}}lcc@{}}
\toprule
\(\lambda_{\rm CRB}\) & ACC & AUROC \\
\midrule
0   & 69.4\(\pm\)1.7 & 82.7\(\pm\)4.2 \\
0.1 & 68.6\(\pm\)1.7 & 84.8\(\pm\)3.6 \\
0.3 & 69.3\(\pm\)1.6 & 84.5\(\pm\)4.9 \\
0.5 & 69.5\(\pm\)1.4 & 84.4\(\pm\)4.5 \\
0.7 & 68.8\(\pm\)1.5 & 85.1\(\pm\)4.0 \\
1 (default) & 69.5\(\pm\)1.7 & 84.0\(\pm\)4.8 \\
\bottomrule
\end{tabular*}
\vspace{-2ex}
\end{table}

\begin{table}[ht]
\centering
\small
\setlength{\tabcolsep}{3pt}
\caption{Sensitivity to the GRS weight under Dirichlet \(\alpha=0.1\).
Dataset headings specify (known/unknown classes, labels per known class).
Weights are multiples of the default; zero disables GRS while retaining CRB.
Entries are single runs with a shared seed within each setting.
ACC and overall AUROC are percentages.}
\label{tab:supp_weight_grs}
\begin{tabular*}{\columnwidth}{@{\extracolsep{\fill}}lcccc@{}}
\toprule
 & \multicolumn{2}{c}{\shortstack{CIFAR-10\\(6/4, 40)}}
 & \multicolumn{2}{c}{\shortstack{CIFAR-100\\(80/20, 40)}} \\
\cmidrule(lr){2-3}\cmidrule(lr){4-5}
\(\lambda_{\rm GRS}\) / default
& ACC & AUROC & ACC & AUROC \\
\midrule
0 (without GRS) & 70.7 & 84.7 & 56.0 & 65.9 \\
0.1            & 71.1 & 84.9 & 56.9 & 69.0 \\
0.3            & 71.3 & 86.9 & 56.8 & 69.5 \\
0.5            & 72.3 & 88.4 & 56.9 & 69.2 \\
1 (default)    & 71.3 & 87.7 & 57.7 & 70.1 \\
3              & 71.5 & 82.0 & 57.7 & 69.5 \\
5              & 71.9 & 83.3 & 56.9 & 66.1 \\
\bottomrule
\end{tabular*}
\vspace{-2ex}
\end{table}


\begin{table}[ht]
\centering
\small
\setlength{\tabcolsep}{3pt}
\caption{CIFAR-10 with 10 labels per known class (60 labeled images); single runs with a shared seed, labeled split, and client partition.
The settings are inherited from the 20-label configuration without re-tuning.
ACC and overall AUROC are percentages; the higher value in each column is bold.}
\label{tab:supp_fewer_labels}
\begin{tabular*}{\columnwidth}{@{\extracolsep{\fill}}lcccc@{}}
\toprule
 & \multicolumn{2}{c}{\(\alpha=0.3\)} & \multicolumn{2}{c}{\(\alpha=0.1\)} \\
\cmidrule(lr){2-3}\cmidrule(lr){4-5}
Method & ACC & AUROC & ACC & AUROC \\
\midrule
FedSSB      & 53.3 & 73.6 & 33.8 & 68.9 \\
\proposal{} & \textbf{55.5} & \textbf{84.1} & \textbf{39.7} & \textbf{69.1} \\
\bottomrule
\end{tabular*}
\end{table}

\begin{table*}[ht]
\centering
\small
\setlength{\tabcolsep}{3pt}
\providecommand{\suppsd}[1]{{\fontsize{7}{7}\selectfont($\pm$#1)}}
\caption{Seen and unseen OOD AUROC (\%) on CIFAR-10 and CIFAR-100.
Each cell shows the mean (top) and standard deviation (bottom) over three seeds, using the same runs and checkpoints as the corresponding main-paper result.
Seen OOD uses unknown classes present in the unlabeled client pools; unseen OOD averages the held-out evaluation datasets.
Their equally weighted mean recovers overall AUROC as in Eq.~\eqref{eq:supp_overall_auc}.}
    \label{tab:supp_ood_breakdown}
    \begin{tabular*}{\textwidth}{@{\extracolsep{\fill}}l*{14}{c}@{}}
    \toprule
    & \multicolumn{2}{c}{SemiFL} & \multicolumn{2}{c}{FedFixMatch} & \multicolumn{2}{c}{FedSCOMatch}
    & \multicolumn{2}{c}{FedProSub} & \multicolumn{2}{c}{FedSSB} & \multicolumn{2}{c}{FedOpenMatch}
    & \multicolumn{2}{c}{\proposal{}} \\
    \cmidrule(lr){2-3}\cmidrule(lr){4-5}\cmidrule(lr){6-7}\cmidrule(lr){8-9}
    \cmidrule(lr){10-11}\cmidrule(lr){12-13}\cmidrule(lr){14-15}
    Partition & Seen & Unseen & Seen & Unseen & Seen & Unseen & Seen & Unseen
    & Seen & Unseen & Seen & Unseen & Seen & Unseen \\
    \midrule
    \multicolumn{15}{c}{\textbf{CIFAR-10 6/4, 20 Labels/Class}} \\
    \midrule
    IID            & 41.9 & 48.7 & 21.5 & 58.7 & 41.9 & 52.7 & 63.0 & 55.6 & 94.1 & 88.9 & 55.1 & 69.2 & 95.0 & 90.6 \\
                   & \suppsd{9.3} & \suppsd{5.5} & \suppsd{12.5} & \suppsd{10.0} & \suppsd{6.8} & \suppsd{3.7} & \suppsd{15.7} &
                   \suppsd{15.0} & \suppsd{2.8} & \suppsd{3.9} & \suppsd{3.8} & \suppsd{12.4} & \suppsd{5.5} & \suppsd{1.2} \\
    $\alpha{=}0.3$ & 48.3 & 50.7 & 19.1 & 63.9 & 40.0 & 53.2 & 49.6 & 46.4 & 78.4 & 80.0 & 35.9 & 59.2 & 93.6 & 81.7 \\
                   & \suppsd{10.1} & \suppsd{9.6} & \suppsd{18.8} & \suppsd{14.2} & \suppsd{5.7} & \suppsd{4.6} & \suppsd{5.0} &
                   \suppsd{5.4} & \suppsd{12.9} & \suppsd{4.4} & \suppsd{7.9} & \suppsd{4.5} & \suppsd{2.8} & \suppsd{3.4} \\
    $\alpha{=}0.1$ & 53.6 & 46.2 & 31.5 & 67.4 & 49.0 & 58.7 & 35.9 & 39.3 & 67.3 & 75.1 & 43.5 & 59.4 & 81.5 & 74.7 \\
                   & \suppsd{4.6} & \suppsd{3.6} & \suppsd{4.4} & \suppsd{4.1} & \suppsd{6.4} & \suppsd{2.8} & \suppsd{13.7} &
                   \suppsd{8.6} & \suppsd{2.9} & \suppsd{0.3} & \suppsd{5.5} & \suppsd{6.1} & \suppsd{5.0} & \suppsd{4.3} \\
    \midrule
    \multicolumn{15}{c}{\textbf{CIFAR-10 6/4, 40 Labels/Class}} \\
    \midrule
    IID            & 26.9 & 49.3 & 36.1 & 77.6 & 46.6 & 58.2 & 80.8 & 69.1 & 92.0 & 91.4 & 58.2 & 61.1 & 90.3 & 89.8 \\
                   & \suppsd{10.4} & \suppsd{18.4} & \suppsd{10.3} & \suppsd{2.4} & \suppsd{5.3} & \suppsd{0.9} & \suppsd{1.1} &
                   \suppsd{2.2} & \suppsd{3.5} & \suppsd{0.9} & \suppsd{12.3} & \suppsd{12.6} & \suppsd{1.1} & \suppsd{0.9} \\
    $\alpha{=}0.3$ & 44.5 & 58.6 & 31.5 & 71.7 & 47.8 & 60.7 & 60.3 & 53.2 & 89.3 & 86.3 & 57.6 & 71.2 & 95.7 & 87.8 \\
                   & \suppsd{5.1} & \suppsd{14.2} & \suppsd{5.2} & \suppsd{5.0} & \suppsd{5.1} & \suppsd{2.6} & \suppsd{2.6} &
                   \suppsd{3.5} & \suppsd{5.0} & \suppsd{2.3} & \suppsd{7.3} & \suppsd{3.8} & \suppsd{1.5} & \suppsd{0.7} \\
    $\alpha{=}0.1$ & 41.4 & 46.8 & 39.9 & 72.2 & 47.1 & 63.2 & 47.3 & 39.8 & 72.9 & 76.9 & 47.7 & 56.7 & 87.8 & 80.3 \\
                   & \suppsd{15.4} & \suppsd{8.5} & \suppsd{3.4} & \suppsd{2.2} & \suppsd{0.9} & \suppsd{6.2} & \suppsd{12.2} &
                   \suppsd{3.8} & \suppsd{2.7} & \suppsd{2.4} & \suppsd{4.6} & \suppsd{11.2} & \suppsd{5.7} & \suppsd{3.9} \\
    \midrule
    \multicolumn{15}{c}{\textbf{CIFAR-100 55/45, 40 Labels/Class}} \\
    \midrule
    IID            & 63.5 & 73.0 & 62.0 & 74.1 & 60.7 & 69.6 & 77.2 & 74.6 & 77.5 & 75.1 & 73.9 & 70.7 & 79.1 & 79.1 \\
                   & \suppsd{0.8} & \suppsd{0.7} & \suppsd{0.7} & \suppsd{2.8} & \suppsd{2.0} & \suppsd{3.8} & \suppsd{0.2} &
                   \suppsd{0.5} & \suppsd{1.5} & \suppsd{3.9} & \suppsd{2.3} & \suppsd{0.5} & \suppsd{1.9} & \suppsd{2.0} \\
    $\alpha{=}0.3$ & 62.3 & 72.9 & 61.2 & 69.5 & 59.5 & 63.4 & 76.6 & 73.9 & 76.0 & 73.7 & 65.6 & 63.0 & 78.1 & 79.3 \\
                   & \suppsd{0.2} & \suppsd{0.6} & \suppsd{0.5} & \suppsd{5.4} & \suppsd{1.3} & \suppsd{1.9} & \suppsd{0.5} &
                   \suppsd{1.4} & \suppsd{2.3} & \suppsd{2.5} & \suppsd{15.6} & \suppsd{12.6} & \suppsd{1.1} & \suppsd{1.6} \\
    $\alpha{=}0.1$ & 61.2 & 69.2 & 61.0 & 70.3 & 60.6 & 68.1 & 71.9 & 71.5 & 74.3 & 71.7 & 50.0 & 50.0 & 77.0 & 75.1 \\
                   & \suppsd{1.0} & \suppsd{2.4} & \suppsd{1.6} & \suppsd{2.7} & \suppsd{1.7} & \suppsd{3.8} & \suppsd{0.3} &
                   \suppsd{1.6} & \suppsd{2.5} & \suppsd{1.3} & \suppsd{0.0} & \suppsd{0.0} & \suppsd{1.9} & \suppsd{2.0} \\
    \midrule
    \multicolumn{15}{c}{\textbf{CIFAR-100 80/20, 40 Labels/Class}} \\
    \midrule
    IID            & 60.1 & 72.0 & 53.8 & 70.6 & 51.5 & 64.2 & 75.0 & 72.6 & 77.7 & 72.5 & 67.2 & 60.4 & 76.6 & 74.4 \\
                   & \suppsd{0.1} & \suppsd{2.4} & \suppsd{0.9} & \suppsd{2.0} & \suppsd{0.8} & \suppsd{4.1} & \suppsd{2.4} &
                   \suppsd{1.4} & \suppsd{1.8} & \suppsd{2.9} & \suppsd{8.0} & \suppsd{3.7} & \suppsd{1.6} & \suppsd{1.3} \\
    $\alpha{=}0.3$ & 58.7 & 70.7 & 54.9 & 66.1 & 51.3 & 63.6 & 71.1 & 72.3 & 73.5 & 71.7 & 69.1 & 60.5 & 75.5 & 72.0 \\
                   & \suppsd{0.8} & \suppsd{1.1} & \suppsd{1.4} & \suppsd{0.3} & \suppsd{1.5} & \suppsd{6.1} & \suppsd{1.5} &
                   \suppsd{0.8} & \suppsd{1.9} & \suppsd{2.9} & \suppsd{1.8} & \suppsd{2.0} & \suppsd{4.1} & \suppsd{2.1} \\
    $\alpha{=}0.1$ & 58.7 & 63.7 & 55.7 & 65.6 & 53.2 & 64.0 & 67.9 & 72.2 & 73.5 & 67.7 & 63.9 & 60.6 & 72.8 & 69.0 \\
                   & \suppsd{0.9} & \suppsd{2.9} & \suppsd{0.5} & \suppsd{5.9} & \suppsd{1.2} & \suppsd{4.7} & \suppsd{2.0} &
                   \suppsd{2.4} & \suppsd{2.5} & \suppsd{1.5} & \suppsd{6.2} & \suppsd{4.7} & \suppsd{2.8} & \suppsd{1.3} \\
    \bottomrule
    \end{tabular*}
    \end{table*}

\subsection{Fewer Labels}
\label{app:fewer_labels}

Table~\ref{tab:supp_fewer_labels} reduces the server-label budget beyond the CIFAR-10 settings of the main paper.
We compare with FedSSB without re-tuning either method, using matched labeled splits, client partitions, and initialization seeds.
In these single-seed runs, \proposal{} achieves higher accuracy under both partitions and higher AUROC under the less heterogeneous partition.
Under stronger heterogeneity, both methods perform worse and their AUROCs are nearly tied.


\subsection{Seen and Unseen OOD Detection}
\label{app:ood_breakdown}

For CIFAR-10 and CIFAR-100, Table~\ref{tab:supp_ood_breakdown} decomposes overall AUROC into seen and unseen OOD performance.
Seen OOD classes occur in the unlabeled client pools.
Unseen evaluation uses the other CIFAR dataset, SVHN, LSUN, ImageNet-30, and Gaussian noise.
Let \(A_{{\rm unseen},m}\) denote AUROC on unseen dataset \(m\).
We use an unweighted mean over the \(M\) unseen datasets and combine it with seen AUROC:
\begin{equation}
\begin{aligned}
A_{\rm unseen}&=\frac{1}{M}\sum_{m=1}^M A_{{\rm unseen},m},\\
A_{\rm overall}&=\frac{1}{2}\big(A_{\rm seen}+A_{\rm unseen}\big).
\end{aligned}
\label{eq:supp_overall_auc}
\end{equation}
The same known-class test images, scoring rule, and evaluation checkpoint are used for both groups.

\proposal{} exceeds SemiFL and FedOpenMatch in mean AUROC for both groups throughout the reported CIFAR settings, and leads both groups in most settings overall.
Exceptions occur on CIFAR-10 and the CIFAR-100 80/20 split, where FedSSB or FedProSub can be stronger.
The improvements therefore extend to unseen datasets, while rankings can differ between OOD groups.